%% file: main.tex
\documentclass[acmsmall,authorversion,nonacm]{acmart}
\usepackage{tabularx}

\usepackage{algorithmic}
\usepackage{graphicx}
\usepackage{textcomp}
\usepackage{xcolor}
\usepackage{booktabs}
\usepackage{tabularx}
\usepackage{multirow}

\usepackage{listings}
\usepackage{multicol}
\usepackage{multirow}
\usepackage{graphicx}
\usepackage{subcaption}
\usepackage{epstopdf}
\usepackage[inline]{enumitem}

\usepackage{amsthm} 
\newtheoremstyle{example}{}{}{}{}{\bfseries}{\smallskip}{\newline}{}
\theoremstyle{example}

\usepackage{todonotes}
\usepackage{tcolorbox}
\usepackage{color, colortbl}
\usepackage{tablefootnote}

\usepackage{soul}
\usepackage{pbox}

\begin{document}

\title{A Review of Bangla Natural Language Processing Tasks and the Utility of Transformer Models}


\author{Firoj Alam}
\email{fialam@hbku.edu.qa}
\orcid{0000-0001-7172-1997}
\affiliation{%
 \institution{Qatar Computing Research Institute, HBKU}
 \city{Doha}
 \country{Qatar}
}

\author{Md. Arid Hasan}
\affiliation{%
 \institution{Cognitive Insight Limited}
 \city{Dhaka}
 \country{Bangladesh}
 }
\email{arid.hasan.h@gmail.com}

\author{Tanvirul Alam}
\affiliation{%
 \institution{BJIT Limited}
 \city{Dhaka}
 \country{Bangladesh}
}
\email{xashru@gmail.com}

\author{Akib Khan}
\affiliation{%
\institution{BJIT Limited}
\city{Dhaka}
\country{Bangladesh}
}
\email{akibkhan0147@gmail.com}

\author{Jannatul Tajrin}
\affiliation{%
 \institution{Cognitive Insight Limited}
 \city{Dhaka}
 \country{Bangladesh}
 }
 \email{jannatultajrin33@gmail.com}

\author{Naira Khan}
\affiliation{%
 \institution{Dhaka University}
 \city{Dhaka}
}
\email{nairakhan@du.ac.bd}

\author{Shammur Absar Chowdhury}
\affiliation{%
 \institution{Qatar Computing Research Institute, HBKU}
 \city{Doha}
 \country{Qatar}
}
\email{shchowdhury@hbku.edu.qa}


\renewcommand{\shortauthors}{Alam F., et al.}

\begin{abstract}

Bangla -- ranked as the $6^{th}$ most widely spoken language across the world,\footnote{\url{https://www.ethnologue.com/guides/ethnologue200}} with 230 million native speakers -- is still considered as a low-resource language in the natural language processing (NLP) community. With three decades of research, Bangla NLP (BNLP) is still lagging behind mainly due to the scarcity of resources and the challenges that come with it. 
There is sparse work in different areas of BNLP; however, a thorough survey reporting previous work and recent advances is yet to be done. In this study, we first provide a 
review of Bangla NLP tasks, resources, and tools available to the research community; we benchmark datasets collected from various platforms for nine NLP  tasks using current state-of-the-art algorithms (i.e., transformer-based models). We provide comparative results for the studied NLP tasks by comparing monolingual vs. multilingual models of varying sizes. We report our results using both individual and consolidated datasets and provide data splits for future research. We reviewed a total of 108 papers and conducted 175 sets of experiments. Our results show promising performance using transformer-based models while highlighting the trade-off with computational costs. We hope that such a comprehensive survey will motivate the community to build on and further advance the research on Bangla NLP.

\end{abstract}



\settopmatter{printacmref=false}
\setcopyright{none}
\renewcommand\footnotetextcopyrightpermission[1]{}
\pagestyle{plain}

\setcopyright{none}
\makeatletter
\renewcommand\@formatdoi[1]{\ignorespaces}
\makeatother

\maketitle

\input{sections/introduction_v2}
\input{sections/background_related_work}
\input{sections/methodology}
\input{sections/experiments}

\input{sections/results}

\input{sections/challenges_future_work}

\input{sections/conclusions}
\bibliographystyle{ACM-Reference-Format}
\bibliography{bib/all_bib}


\end{document}

%% file: sections/introduction_v2.tex
\section{Introduction}
\label{sec:intro}

Bangla is one of the most widely spoken languages in the world~\cite{ali2010unsupervised}, with nearly 230 million native speakers. The language is morphologically rich with diverse dialects and a long-standing literary tradition that developed over the course of thousands of years. Bangla is also the only language that has inspired a language movement, which gained international recognition and is now celebrated as International Mother Language Day.\footnote{The day of the movement -- $21^{st}$ February is now recognized by UNESCO as International Mother Language Day} However, Bangla is still considered a low-resource language in terms of digitization, primarily due to the scarcity of annotated computer readable datasets and the limited support for resource building. 

Research on Bangla Natural Language Processing (BNLP) began in the early 1990s, with an initial focus on rule-based lexical and morphological analysis \cite{islam2009research,sengupta1993lexical,sengupta1993morpho}. 
Later the advancement in BNLP research continued in the 2000s, with a particular focus on Parts of Speech (POS) tagging \cite{hasan2007comparison}, grammar checkers \cite{alam2007n}, Named Entity Recognition (NER) \cite{ekbal2010named,alam2016bidirectional}, morphological analysis \cite{dasgupta2005morphological,dasgupta2004morphological}, parsing using context free grammars\cite{mosaddeque2004context}, Machine Translation (MT)~\cite{sennrich2015neural}, Text-to-Speech (TTS) \cite{alam2007text,alam2008text}, Speech Recognition \cite{hasnat2007isolated,chowdhury2010implementation,paul2009bangla}, and Optical Character Recognition (OCR)~\cite{chaudhuri1998complete,hasnat2008high,hasnat2009open}. 
Over the years, the research focus has extended to include automated text summarization, sentiment analysis \cite{hasan2013sentiment}, emotion detection~\cite{das2010sentiwordnet}, and news categorization~\cite{article}.


Feature engineering has been ubiquitous in earlier days. For text classifications, the features such as token, bag-of-words, and bag-of-n-grams are used to feature representation. In sequence tagging, hand-crafted features include mainly orthographic features such as prefix, suffix, the word in context, abbreviation, and numbers. Task-specific knowledge source such as the lexicon has also been used, which lists whether a word is a noun, pronoun or belongs to some others classes \cite{alam2016bidirectional}.

From the modeling perspective, most of the earlier endeavors are either rule-based, statistical or classical machine learning based approaches. Classical machine algorithms include Naive Bayes (NB), Support Vector Machine (SVM) \cite{kudo2001chunking,yamada2003statistical}, and Random Forests (RF)~\cite{breiman2001random} are used for the text classification task. As for the sequence tagging tasks, such as NER and G2P, the algorithms include Hidden Markov Models (HMMs) \cite{brants2000tnt}, Conditional Random Fields (CRFs) \cite{lafferty2001conditional}, Maximum Entropy (ME) \cite{ratnaparkhi1996maximum}, Maximum Entropy Markov Models (MEMMs) \cite{mccallum2000maximum}, and  hybrid approach \cite{altun2003hidden}.

It is only very recently that a small number of studies have explored deep learning-based approaches \cite{arid2019MTb,AridSentiment2020,alam2020bangla,bhattacharjee2021banglabert}, which include Long Short Term Memory (LSTM) neural networks \cite{hochreiter1997long} and Gated Recurrent Unit (GRU) \cite{cho2014learning} and a combination of LSTM, Convolution Neural Networks (CNN) \cite{lecun1989backpropagation} and CRFs \cite{huang2015bidirectional,ma2016end,lample2016neural}. Typically, these algorithms are used with distributed words and character representations called ``word embeddings" and ``character embeddings," respectively.



For a low-resource language like Bangla, resource development and benchmarks have been major challenges. At present, publicly available resources are limited 
and are focused on certain sets of annotation like  sentiment \cite{tripto2018detecting,ashik2019data}, news categorization data \cite{kunchukuttan2020ai4bharat}, authorship attribution data \cite{khatun2019authorship}, speech corpora \cite{alam2010development,alam2008consacoustic}, parallel corpora for machine translation \cite{hasan2020collaborative,hasan2019icbslp, hasan2019neural}, and pronunciation lexicons \cite{alam2008text,chowdhury2017bangla}. A small number of recent endeavors report benchmarking sentiment and text classification tasks \cite{AridSentiment2020,alam2020bangla}.


Previous attempts to summarize the contributions in BNLP includes a book (released in 2013) \cite{karim2013technical} highlighting different aspects of Bangla language processing, complied addressing the limitations as well as the progress made at the time. 
The book covered contents including font design, machine translation, character recognition, parsing, a few avenues of speech processing, information retrieval, and sentiment analysis. 
Most recently, a survey on Bangla language processing reviews the work appeared from 1999 to 2021, where eleven different topics have been addressed in terms of classical to deep learning approaches \cite{sen2021bangla}. 

In this work, we aim to provide a comprehensive survey on the most notable NLP tasks addressed for Bangla, which can help the community with a direction for future research and advance further study in the field.
Unlike the above mentioned survey works, in this study, we conduct extensive experiments using nine different transformer models to create benchmarks for nine different tasks. We also conduct extensive reviews on these tasks to compare our results, where possible. There has not been any survey on Bangla NLP tasks of this nature; hence this will serve as a first study for the research community. 
Since covering the entire field of Bangla NLP is difficult, we focused on the following tasks: {\em(i)} POS tagging, {\em(ii)} lemmatization, {\em(iii)} NER, {\em(iv)} punctuation restoration, {\em(v)} MT, {\em(vi)} sentiment classification, {\em(vii)} emotion classification, {\em(viii)} authorship attribution, and {\em(ix)} news categorization. 



Our contributions in the current study are as follows: 
\begin{enumerate}
    \itemsep0em
    \item We provide a detailed survey on the most notable NLP tasks by reviewing 108 papers.
    \item We benchmark different (nine) tasks with experimental results using nine different transformer models,\footnote{For MT we use one pre-trained transformer model.} which resulted in 175 sets of experiments.
    \item We provide comparative results for different transformer models comparing (1) models' size (large {\em vs.} small) and (2) style (mono {\em vs.} multilingual models).
    \item We also report comparative results for individual vs. consolidated datasets, when multiple data source is available. 
    \item We analyze the trade-off between performance and computational complexity between the transformer-based and classical approaches like SVM.
    \item We provide a concrete future direction for the community answering to questions like: (1) what resources are available? (2) the challenges? and (3) what can be done?
    \item We provide data splits for reproducibility and future research.\footnote{\url{https://github.com/banglanlp/bnlp-resources} Note that we could only provide and share data splits, which were publicly accessible. Any private data can be possibly accessed by contacting respective authors.}  
\end{enumerate}



The rest of the paper is structured as follows: In Section \ref{sec:background}, we provide a background study of different approaches to BNLP, including the research conducted to date. We then describe the pre-trained transformer-based language models Section \ref{sec:methods}. We proceed to discuss the data and experiments in each area of BNLP in Section \ref{sec:experiments}. We also report the experimental results and discuss our findings Section \ref{sec:results}. The individual overall performance, limitation, and future work are discussed in Section \ref{sec:challenge_future_work}. Finally, we conclude our study with the future direction of BNLP research in Section \ref{sec:conclusion}.

%% file: sections/background_related_work.tex
\section{Background and Related Work}
\label{sec:background}
As we reviewed the literature, we delved into past work with a focus on particular topics, resulting in a literature review that covers the last several decades. 
Our motivation for such an extensive exploration is manifold: {\em (i)} As Bangla is a low-resource language and there is a dearth of NLP related research, we aimed to search through previous work to see if there are past resources developed in the early days that can be pushed forward, {\em (ii)} providing the community a heads-start, along with a sense of past and present conditions,
 {\em (iii)} providing a future research direction and {\em (iv)} benchmarks to build on.

\subsection{Parts of Speech}
\label{ssec:pos_related_work}

\begin{table}[tbh!]
\centering
\caption{Relevant work in the literature for \textbf{POS tagging}. Reported results are in Accuracy (Acc)}
\label{tab:pos-table}
\setlength{\tabcolsep}{3pt}
\scalebox{0.8}{
\begin{tabularx}{\textwidth}{@{}XXXX@{}}
\toprule
\multicolumn{1}{c}{\textbf{Paper}} &
\multicolumn{1}{c}{\textbf{Technique}} &
\multicolumn{1}{c}{\textbf{Datasets}} &
\multicolumn{1}{c}{\textbf{Results (Acc)}} \\ \hline
Dandapat et al. \cite{dandapat2004hybrid} & HMM & 500 tagged sentences and 50,000 untagged words for training, CIIL corpus for test & 95.0\% \\
Hasan et al. \cite{hasan2007comparison} & HMM, Unigram, and Brill & 5,000 words from Prothom Alo & 45.6\%, 71.2\%, and 71.3\% using 12 TAGs and 46.9\%, 42.2\%, and 54.9\% using 41 TAGs \\
Ekbal et al. \cite{ekbal2007bengali} & CRF, NER, Lexicon, UNK word features & NLPAI\_Contest06 and SPSAL2007 contest data & 90.3\% \\
Ekbal et al. \cite{ekbal2008part} & HMM, ME, CRF, SVM & NLPAI\_Contest06 and SPSAL2007 contest data & HMM: 78.6\%, ME: 83.3\%, CRF: 85.6\%, SVM: 86.8\% \\
Ekbal et al. \cite{ekbal2008web} & HMM, SVM & Bengali news corpus & HMM: 85.6\%, SVM: 91.2\% \\
Dhandapat et al. \cite{dandapat2007automatic} & HMM (Supervised and Semi-supervised), ME & 3,625 tagged and 11,000 untagged sentences from EMILLE corpus & HMM-S: 88.7\%, HMM-SS: 87.9\%, and ME: 88.4\% \\
Ekbal et al. \cite{ekbal2008maximum} & ME & NLPAI\_Contest06 and SPSAL2007 & 88.2\% \\
Dandapat et al. \cite{dandapat2006part} & HMM (Supervised and Semi-supervised) & NLPAI and 100,000 unlabeled tokens & HMM-S + CMA: 88.8\%, HMM-SS + CMA: 89.6\% \\
Sarker et al. \cite{sarkar2012practical} & HMM & N/A, NLTK data for test & 78.7\% \\
Mukherjee et al. \cite{mukherjee2013bengali} & Global Linear Model & SPSAL2007 & 93.1 \% \\
Ghosh et al. \cite{ghosh2016part} & CRF & ICON-2015 & Bengali-English: 75.2\% \\
Ekbal et al. \cite{ekbal2009voted} & ME, HMM, CRF & NLPAI\_Contest06 and SPSAL2007 & ME: 81.9\%, SVM: 85.9\%, CRF: 84.2\% \\
Kabir et al. \cite{kabir2016deep} & Deep Learning & IL-POST project & 93.3\% \\
Alam et al. \cite{alam2016bidirectional} & BiLSTM-CRF & LDC corpus & 86.0\% \\
Hoque et al. \cite{hoque2015bangla} & Rule Based & self developed & 93.7\% \\
\bottomrule
\end{tabularx}%
}
\end{table}

Parts-of-Speech (POS) tagging plays a key role in many areas of linguistic research \cite{schmid1994part, chakrabarti2011layered}. The advent of POS tagging research for Bangla can be traced back to the early 2000s \cite{dandapat2004hybrid}, which includes the study of rule-based systems \cite{dandapat2004hybrid}, statistical models \cite{dandapat2004hybrid, ekbal2007bengali, hasan2007comparison, ekbal2008part}, and unsupervised models \cite{ali2010unsupervised}. In Table \ref{tab:pos-table}, we provide a brief overview of the relevant studies, with the associated datasets, methodologies, and respective results. 

The study of Dandapat et al. \cite{dandapat2004hybrid} reports an HMM-based semi-supervised approach with the use of 500 tagged and 50,000 untagged sentences, which achieved an accuracy of 95\% on 100 randomly selected sentences from the CIIL corpus. In \cite{hasan2007comparison}, the authors report a comparative study among different approaches, which includes Unigram, HMM, and Brill's approaches on 5,000 Bangla words, taken from the Prothom Alo newspaper. The authors noted the following levels of accuracy on the two sets: {\em (i)} using 12 tags, 45.6\% with HMM, 71.2\% with Unigram, 71.3\% with Brill's approach, {\em (ii)} using 41 tags, 46.9\% with HMM, 42.2\% with Unigram, 54.9\% using Brill's approach. The study of Automatic Part-of-Speech tagging was conducted by Dandapat et al. \cite{dandapat2007automatic}. The authors report supervised and semi-supervised HMM and ME-based approaches on EMILLIE/CIIL corpus in the said study. The study reports 88.75\%, 87.95\%, and 88.41\% for HMM supervised, HMM semi-supervised, and ME, respectively, using suffix information and a morphological analyzer. In another study \cite{dandapat2006part}, Dandapat et al. report HMM-based supervised and semi-supervised approaches with the use of a morphological analyzer on NLPAI contest data and 100,000 unlabeled tokens. The study reports accuracy levels comprising 88.83\% and 89.65\% for HMM-based supervised and semi-supervised approaches, respectively.

Ekbal et al. had a series of studies for POS taggers \cite{ekbal2007bengali,ekbal2008part,ekbal2008web,ekbal2008maximum}.  In \cite{ekbal2007bengali}, Ekbal et al. reported a method that combines NER, lexicon and unknown word features with CRF. The study used NLPAI\_Contest06 and SPSAL2007 contest data to evaluate a CRF-based POS tagger and achieved an accuracy of 90.30\%. In another study \cite{ekbal2008part} Ekbal et al. reports an SVM based approach for POS tagging. Additionally, the authors used HMM, ME, and CRF approaches to compare with the proposed SVM-based approach. The study reports an accuracy of  86.84\% for the proposed model, which outperforms existing approaches. In \cite{ekbal2008web}, the authors report an HMM and SVM-based Bangla POS tagger with the use of NER and lexicon and handling unknown word features on Bangla News corpus. The study reports an accuracy of 85.56\% and 91.23\% using HMM and SVM, respectively. In another study \cite{ekbal2008maximum}, Ekbal et al. explore a ME-based Bangla POS tagger on NLPAI\_Contest06 and SPSAL2007 contest data. With ME, the said study utilizes lexical resources, NER inflections, and unknown word handling features, which achieved an accuracy of 88.20\%. In \cite{ekbal2009voted}, Ekbal et al. proposed a voted approach using NLPAI\_contest06 and SPSAL2007 workshop data and achieved an accuracy of 92.35\%. In the study, they used ME, HMM, and CRF to compare with the proposed model.

The study of Sarkar et al. \cite{sarkar2012practical} reports a POS tagging system using an HMM-based approach and achieved an accuracy of 78.68\% using trigrams and HMMs. In \cite{mukherjee2013bengali}, the authors proposed a Global Linear Model on the SPSAL 2007 workshop data and achieved an accuracy of 93.12\%. The study also executed CRF, SVM, HMM, and ME-based POS taggers to compare with the proposed model.

The work of Gosh et al. \cite{ghosh2016part} was in a considerably different direction. Their study code-mixed social media text using CRF, in which they achieved an accuracy of 75.22\% on the Bengali-English of the 12th International Conference on Natural Language Processing shared task.

 In \cite{kabir2016deep}, the authors report a deep learning-based POS tagger on Microsoft Research India as part of the Indian Language Part-of-Speech Tagset project. Using Deep Belief Network for training and evaluation, they achieved an accuracy of 93.33\%. 
 
Hoque et al. in \cite{hoque2015bangla}, report a stemmer and rule-based analyzer for Bangla POS tagging and achieved an accuracy of 93.70\%. In a recent work, Alam et al. \cite{alam2016bidirectional} reports Bidirectional LSTMs-CRFs networks for Bangla POS tagging on the LDC corpus developed by Microsoft Research India and achieved 86\% accuracy.


\subsection{Stemming and Lemmatization}
\label{ssec:lemmatization_related_work}

\begin{table}[t]
\centering
\caption{Relevant work in the literature for \textbf{stemmer and lemmatization}. Reported results are in Accuracy (Acc). FFNN: Feed-Forward Neural Network. P: Precision. MAP: Mean Average Precision}
\label{tab:stemma-table}
\setlength{\tabcolsep}{3pt}
\scalebox{0.8}{
\begin{tabularx}{\textwidth}{@{}lXXX@{}}
\toprule
\multicolumn{1}{c}{\textbf{Paper}} &
\multicolumn{1}{c}{\textbf{Technique}} &
\multicolumn{1}{c}{\textbf{Datasets}} &
\multicolumn{1}{c}{\textbf{Results (Acc)}} \\ \hline
Majumder et al. \cite{majumder2007yass} & Suffix striping based & 50,000 News documents & P 49.6 \\ 
Urmi et al. \cite{urmi2016corpus} & N-gram & Bangla corpus from different online sources & 40.2 \% \\
Sarker et al. \cite{sarkar2012fire} & Rule-based & CLC and CTC & CLC: 96.4\%, CTC: 92.6\%, and overall: 94.7\% \\
Dolamic et al. \cite{dolamic2008unine} & 4-gram and light stemmer & News from CRI and Anandabazar Patrika\footnote{\url{https://www.anandabazar.com/}} & light: 41.3\% and 4-gram: 40.7\% \\
Seddiqui et al. \cite{seddiqui2016recursive} & Recursive suffix striping & Prothom Alo\footnote{\url{https://www.prothomalo.com/}} (0.78 million words) & 92.0\% \\
Paik et al. \cite{paik2008simple} & TERRIER & FIRE-2008 & 42.3\% \\
Ganguly et al. \cite{ganguly2012dcu} & Rule-based & FIRE-2011 & 33.0\% \\
Das et al. \cite{das2010morphological} & K-means clustering & Project IL-ILMT & 74.6\% \\
Das et al. \cite{das2011rule} & Rule-based & FIRE-2010 & 47.5\% \\
Sarker et al. \cite{sarkar2008design} & Rule-based & 3 short stories of Rabindranath Tagore & RT1: 98.8\%, RT2: 98.7\%, and RT3: 99.9\% \\
Islam et al. \cite{islam2007light} & Suffix striping & N/A, 13,000 words for test & Single error acc: 90.8\%, multi-error acc: $\sim$67.0\% \\
Mahmud et al. \cite{mahmud2014rule} & Rule-based & Prothom Alo and BDNews24\footnote{\url{https://bangla.bdnews24.com/}} articles & Verb: 83.0\%, Noun: 88.0\% \\
Chakrabarty et al. \cite{chakrabarty2016neural} & FFNN & 19,159 training and 2,126 words test  & 69.6\% \\
Loponen et al. \cite{loponen2013uta} & Stale, Yass and Grale & FIRE-2010 & MAP 54.4\% \\
Chakrabarty et al. \cite{chakrabarty2017context} & BiLSTM, BiGRU & Tagore’s stories and articles Anandabazar Patrika & BiLSTM: 91.1\% and BiGRU: 90.8\% \\
Chakrabarty et al. \cite{chakrabarty2016benlem} &  & 18 articles from FIRE Bengali News Corpus & 81.9\% \\
Pal et al. \cite{pal2015innovative} & Longest suffix stripping & Pashchimbanga Bangla Akademi\footnote{\url{http://banglaacademy.org.in/bangla_academy.html}} & 94.0\% \\
\bottomrule
\end{tabularx}%
}
\end{table}

Stemming is the process of removing morphological variants of a word to map it to its root or stem~\cite{martin2000speech}. The process of mapping a wordform to a lemma\footnote{``A lemma is a set of lexical forms having the same stem, the same major part-of-speech, and the same word-sense. The wordform is the full inflected or derived form of the word.'' \cite{martin2000speech}} is called lemmatization~\cite{martin2000speech}. 
\subsubsection{Stemming:}
Like other BNLP areas of research, the work on Bangla stemming started a little over a decade ago \cite{islam2007light,paik2008simple,dolamic2008unine}. 
The study of Islam et al. \cite{islam2007light} reports a lightweight stemmer for a Bangla spellchecker. The authors used as a resource a 600 root word lexicon and a list of 100 suffixes. The system was tested using 13,000 words and achieved a single error accuracy of 90.8\% and a multi error accuracy of $\sim$67\%. 

The study of Paik et al. \cite{paik2008simple} reports a simple stemmer using the TERRIER model for indexing and retrieval of information. In the study, the authors report a MAP score of 43.32\% on the FIRE 2008 dataset. A light stemmer and a 4-gram stemmer have been studied by Dolamic et al. \cite{dolamic2008unine}, which reports an accuracy of 41.31\% and 40.74\%, in the light and the 4-gram stemmer, respectively, using news data from September 2004 to September 2007 of CRI and Anandabazar Patrika. In addition to the supervised approach, clustering approaches have also been explored for stemming. In \cite{das2010morphological}, the authors used K-means clustering on the IL-ILMT dataset and reported an accuracy of 74.6\%. 

Apart from the supervised and semi-supervised approaches, earlier work also includes rule-based approaches. Das et al. \cite{das2011rule} proposed a rule-based stemmer with the FIRE 2010 dataset, in which they report a MAP score of 47.48\%. In another study \cite{sarkar2012fire}, the authors proposed Mulaadhaar -- a rule-based Bengali stemmer with the FIRE 2012 task. The proposed approach consists of the use of two corpora, such as the Classic Literature Corpus (CLC) with 15,347 tokens and the Contemporary Travelogue Corpus (CTC) with 11,561 tokens, as well as their combination. The accuracy of their systems is 96.4\%, 92.6\%, and 94.7\% on CLC, CTC, and the combined corpus, respectively. In \cite{ganguly2012dcu}, the authors report a rule-based stemmer using the FIRE 2011 document collection dataset that achieved an accuracy of 33\% for Bangla, which is the second-best result in the MET-2012 task. Another rule-based approach proposed by Mahmud et al. \cite{mahmud2014rule}, used data from Prothom Alo\footnote{\url{https://www.prothomalo.com/}} and BDNews24\footnote{\url{https://bangla.bdnews24.com/}} newspapers, and achieved an accuracy of 88\% for verbs and accuracy of 83\% for nouns.
In \cite{seddiqui2016recursive}, the authors propose a recursive suffix stripping approach for stemming Bangla words. In the study, the authors collected 0.78 million words from the Prothom Alo newspaper and achieved an overall accuracy of 92\% on this dataset. In \cite{urmi2016corpus}, the authors report a corpus-based unsupervised approach using an N-gram language model on Bangla data, which was collected from several online sources. The authors used a 6-gram model and achieved an accuracy of 40.18\%.

\subsubsection{Lemmatization:}
Some of the earlier work on lemmatization of Bangla can be found in the study of Majumder et al. \cite{majumder2007yass}. It proposed a string distance-based approach for Bangla word-stemming on 50,000 news documents and achieved an accuracy of 49.60\% (i.e., 39.3\% improvement over the baseline result).  
In \cite{loponen2013uta}, the authors proposed a lemmatizer with three language normalizers (YASS stemmer, GRALE lemmatizer, and StaLe lemmatizer) on the FIRE 2010 dataset. The study also reports that the query with title-description-narrative provides the best MAP accuracy i.e., 54.38\%.
The study of Pal et al. \cite{pal2015innovative} proposed the longest suffix-stripping based approach using a  wordlist collected from the Pashchimbanga Bangla Akademi, Kolkata. The authors achieved a 94\% accuracy on the wordlist. In \cite{chakrabarty2016benlem}, the authors proposed a rule-based Bangla lemmatizer used on 18 random news articles of the FIRE Bengali News Corpus consisting of 3,342 surface words (excluding proper nouns). The authors reported accuracy of 81.95\%.

Rule-based approaches have been dominant for Bangla stemmers and lemmatizers. However, recently deep learning-based approaches have been explored in a number of studies. Chakrabarty et al. \cite{chakrabarty2016neural} proposed a feed-forward neural network approach; trained and evaluated their system on a corpus of 19,159 training samples and 2,126 test samples. The reported accuracy of their system is 69.57\%. In another study \cite{chakrabarty2017context}, the authors report a context-sensitive lemmatizer using two successive variant neural networks. The authors used Tagore stories and news articles from Anandabazar Patrika to train and evaluate the networks. The authors report on BiLSTM and BiGRU networks and achieved maximum accuracies of 91.14\% and 90.85\% for BiLSTM-BiLSTM and BiGRU-BiGRU, respectively, with restricting output classes.

In Table \ref{tab:stemma-table}, we provide relevant work and the corresponding techniques used, datasets, and results of each study.

\subsection{Named Entity Recognition}
\label{ssec:ner_related_work}

\begin{table}[t]
\centering
\caption{Relevant work in the literature for \textbf{NER}. Results are mostly reported in F1 score. For a few work other metrics has been used, which are mentioned.}
\label{tab:ner-table}
\setlength{\tabcolsep}{3pt}
\scalebox{0.8}{
\begin{tabularx}{\textwidth}{@{}XXXX@{}}
\toprule
\multicolumn{1}{c}{\textbf{Paper}} &
\multicolumn{1}{c}{\textbf{Technique}} &
\multicolumn{1}{c}{\textbf{Datasets}} &
\multicolumn{1}{c}{\textbf{Results (F1)}} \\ \hline
Ekbal et al. \cite{ekbal2010named} & SVM & IJCNLP-08 NER shared task & 84.1\% \\
Ekbal et al. \cite{ekbal2009bengali} & Combination of ME, CRF, and SVM & 150K wordforms collected from newspaper &  85.3\% \\
Ekbal et al. \cite{ekbal2009named} & Voted System & 22K wordforms of the IJCNLP-08 NER Shared Task & Acc 92.3\% \\
Ekbal et al.\cite{ekbal2008web} & NER system with linguistic features & Training sentences 44,432 and test sentences 5,000 & 75.4\%, 72.3\%, 71.4\%, and 70.1\% for person, location, organization, and miscellaneous names, respectively 
\\
Ekbal et al. \cite{ekbal2008development} & SVM & Sixteen-NE
tagged corpus of 150K wordforms & 91.8\% \\
Ekbal et al. \cite{ekbal2008bengali} & SVM & 150K words & 91.8\% \\
Ekbal et al. \cite{ekbal2008named} & CRF & 150K words & 90.7\% \\
Ekbal et al. \cite{ekbal2007hidden} & HMM & 150k wordforms & 84.5\% \\
Banerjee et al. \cite{banerjee2014bengali} & Margin Infused Relaxed Algorithm & IJCNLP-08 NERSSEAL & 89.7\% \\
Hasanuzzaman et al. \cite{hasanuzzaman2009maximum} & Maximum Entropy & IJCNLP-08 NER Shared Task & Acc 85.2\% \\
Singh et al. \cite{singh2008named} & N/A(Shared Task) & IJCNLP-08 NER Shared Task & 65.9\% \\
Chaudhuri et al. \cite{chaudhuri2008experiment} & Combination of dictionary-based, rule-based and n-gram based approaches & 20,000 words in training set & 89.5\% \\
Hasan et al. \cite{hasan2009learning} & Learning-based & 77,942 words & Acc 72.0\% \\
Chowdhury et al. \cite{chowdhury2018towards} & CRF & Bangla Content Annotation Bank & 58.0\% \\
\bottomrule
\end{tabularx}%
}
\end{table}

The work related to Bangla NER is relatively sparse. The current state-of-the-art for Bangla NER shows that most of the work has been done by Ekbal et al. \cite{ekbal2010named,ekbal2009bengali,ekbal2009named,ekbal2008web,ekbal2008development,ekbal2008bengali,ekbal2008named,ekbal2007hidden} and IJCNLP-08 NER Shared Task \cite{singh2008named}. Studies by Ekbal et al. comprise the NER corpus development, feature engineering, the use of HMMs, SVMs, ME, CRFs, and a combination of classifiers. The reported F1 measure varies from 82\% to 91\% across a corpus with a different number of entity types. The study of Chaudhuri et al. \cite{chaudhuri2008experiment} used a hybrid approach, which includes a dictionary and rule and n-gram based statistical modeling. 

The study in \cite{hasan2009learning} focused on the geographical context of Bangladesh, as they collected and used data from one of the Bangladeshi Newspapers, namely Prothom-Alo \cite{hasanuzzaman2009maximum}. In their study, only three entity types (i.e., tags) are annotated i.e. \textit{person, location} and \textit{organization}. The reported accuracy of their study is F1 71.99\%. 
Banerjee et al. proposed the Margin Infused Relaxed Algorithm for NER, where they used the IJCNLP-08 NERSSEAL dataset for the experiment \cite{banerjee2014bengali}. In \cite{ibtehaz2018partial}, Ibtehaz et al. reported a partial string matching technique for Bangla NER.

In \cite{chowdhury2018towards}, Chowdhury et al. developed a corpus for NER consisting of seven entity types. The entities are annotated in different newspaper articles collected from various Bangladeshi newspapers. The study investigates token, POS, gazetteers, contextual features, and conditional random fields (CRFs) and utilizes BiLSTM with CRF network. In this study, we used the same corpus and utilized transformer models to provide a new benchmark. 
In Table \ref{tab:ner-table}, we provide a list of the recent work done on NER tasks for Bangla. 

\subsection{Punctuation Restoration}
\label{ssec:punc_related_work}

Punctuation restoration is the task of adding punctuation symbols to raw text. It is a crucial post-processing step for punctuation-free texts that are usually generated using Automatic Speech Recognition (ASR) systems. It makes the transcribed texts easier to understand for human readers and improves the performance of downstream NLP tasks. State-of-the-art NLP models are usually trained using punctuated texts (e.g., texts from newspaper articles, Wikipedia); hence the lack of punctuation significantly degrades performance. As an example, for a Named Entity Recognition system, there is a performance difference of more than $\sim10\%$ when the model is trained with newspaper texts and tested with transcriptions as reported in \cite{alam2015comparing}.

Most of the earlier work on punctuation restoration has been done using lexical, acoustic, and prosodic features or a combination of these features \cite{gravano2009restoring,levy2012effect,zhang2013punctuation,xu2014deep,szaszak2019leveraging,che2016sentence}. Lexical features are widely used for the task as the model can be trained with any well-punctuated text that is readily available, e.g., newspaper articles, Wikipedia, etc. In terms of machine learning models, Conditional Random Fields (CRFs) have been widely used in earlier studies \cite{lu2010better,zhang2013punctuation}. Lately, the use of deep learning models, such as Long Short-Term Memory (LSTM), Convolutional Neural Network (CNN), and transformers have also been used \cite{DBLP:conf/lrec/CheWYM16,gale2017experiments,DBLP:conf/interspeech/ZelaskoSMSCD18,wang2018self} for the task of punctuation restoration.

The only notable work on Bangla punctuation restoration was done in \cite{alam-etal-2020-punctuation}. The authors explored different Transformer architectures for punctuation restoration in English and Bangla. For Bangla, they used different multi-lingual transformer models, including multi-lingual BERT \cite{devlin2018bert}, XLM \cite{xlm}, and XLM-RoBERTa \cite{xlm-roberta}. Using XLM-RoBERTa and data augmentation, they obtained a 69.5 F1-score on the manually transcribed texts and a 64.0 F1-score on the ASR transcribed texts. 


\subsection{Machine Translation}
\label{ssec:mt_related_work}

\begin{table}[t]
\centering
\caption{Relevant work in the literature for \textbf{MT}. $*$ No mention of dataset and results, $\ddagger$ survey paper. Results are mostly reported in BLUE scores. For a few work other metrics has been used, which are mentioned.}
\label{tab:mt-table}
\setlength{\tabcolsep}{4pt}
\scalebox{0.9}{
\begin{tabularx}{\textwidth}{lXXX}
\toprule
\multicolumn{1}{c}{\textbf{Paper}} &
\multicolumn{1}{c}{\textbf{Technique}} &
\multicolumn{1}{c}{\textbf{Datasets}} &
\multicolumn{1}{c}{\textbf{Results (BLUE)}} \\ \hline
Naskar et al. \cite{naskar2006handling}$*$ $\ddagger$ & Rule based & N/A & N/A \\
Salam et al. \cite{salm2009example}$*$ & Example-based & N/A & N/A \\
Fransisca et al. \cite{francisca2011adapting} & Adapting rule based & 79 and 27 sentences for training and test set & 25 correct sentences \\
Dasgupta et al. \cite{dasgupta2004optimal}$*$ & Rule based & N/A & N/A \\
Chatterji et al. \cite{chatterji2009hybrid} & Lexical transfer based SMT & EMILLE-CIIL & 22.7 \\
Saha et al. \cite{saha2005semantics} & Example-based & 2000 News Headlines & N/A \\
Adak et al. \cite{adak2013advanced} & Rule-based & Most common Bengali words & F1 81.5\% \\
Antony et al. \cite{antony2013machine}$*$ & N/A & N/A & N/A \\
Naskar et al. \cite{naskar2005use}$*$ & N/A & N/A & N/A \\
Islam et al. \cite{islam2010english} & Phrase-based SMT & EMILLE-CIIL & Overall 11.7, short sentences 23.3 \\
Pal et al. \cite{pal2011handling} & Phrase-based SMT & EILMT System & 15.1 \\
Ashrafi et al. \cite{ashrafi2013english}$*$ & CFG & N/A & N/A \\
Banerjee et al. \cite{banerjee2018multilingual} & Many-to-One Phrase-based SMT & ILMPC & 14.0 \\
Haffari et al. \cite{haffari2009active} & Active learning based SMT & Hansards and EMILLE & 5.7 \\
Mumin et al. \cite{mumin2019} & Log-linear phrase based SMT & SUPara & 17.4 \\
Dhandapat et al. \cite{dandapat2018training} & Vanilla phrase-based SMT and NMT & Microsoft dataset & 16.6 \& 20.2 \\
Khan et al. \cite{khan2017machine} & Phrase-based SMT & EMILLE & 11.8 \\
Mishra et al. \cite{mishra2014shata} & Phrase-based SMT & SATA-Anuvadok & 18.2 \\
Post et al. \cite{post2012constructing} & Phrase-based SMT & SIPC & 12.7 \\
Liu et al. \cite{liu2018augmenting} & Seq2Seq & N/A & 10.9 \\

Hasan et al. \cite{hasan2019icbslp} & SMT, BiLSTM & ILMPC, SIPC, PTB & 14.8 \& $15.6$\\
Hasan et al. \cite{hasan2019icbslp} & BiLSTM & SUPara & 19.8 \\
Hasan et al. \cite{hasan2019neural} & BiLSTM, Transformer & ILMPC, SIPC, PTB & 15.6 \& 16.6 \\
Hasan et al. \cite{hasan2019neural} & BiLSTM & SUPara & 20.0\\
Mumin et al. \cite{mumin2019neural} & NMT & SUPara & 22.7\\
Hasan et al. \cite{hasan2020not} & Transformer & 2.75M data consolidated from different corpora and websites & 32.1\\
\bottomrule
\end{tabularx}%
}
\end{table}

From the advent of Machine Translation (MT), rule-based systems have been extensively studied, with statistical approaches being introduced in the 1980s \cite{brown1988statistical}. Since then, MT has undergone rapid advancement, with neural networks resulting in state-of-the-art performance. Initial studies in MT for the Bangla-English language pair can be traced to the early 1990s \cite{naskar2005use}. The study of Naskar et al. \cite{naskar2004anubaad} reports the first Bangla-English MT system using a rule-based model.
In \cite{naskar2006handling}, the authors proposed an MT system by analyzing prepositions for both Bangla and English. In said study, the authors claimed that the prepositional word choice depends on the semantic information to translate from English to Bangla. In another study \cite{salm2009example}, the authors proposed an example-based MT system using WordNet for the Bangla-English language pair. A rule-based approach has been adapted by Francisca et al. \cite{francisca2011adapting} in which they developed a set of predefined rules by examining sentence structures. In the study above, the system correctly generated the translation of 25 out of 27 sentences. Another rule-based approach has been proposed by Dasgupta et al. \cite{dasgupta2004optimal} in which rules from source sentences were extracted using a parse tree, with the parse tree then transferred to the target sentence rules.

Statistical approaches have been studied for the Bangla-Hindi language pair by Chatterji et al. \cite{chatterji2009hybrid}. In the said study, the authors used the EMILLIE-CILL parallel corpus to train and evaluate the lexical transfer-based statistical machine translation (SMT) and achieved a BLEU score of 22.75. The study of Saha et al. \cite{saha2005semantics} reports an example-based MT system by analyzing semantic meanings on 2000 news headlines from The Statesman newspaper. In \cite{adak2013advanced}, the authors proposed an advanced rule-based approach for a Bangla-English MT system. In the said study, sentences were analyzed using a POS tagger and then matched with rules, the most common Bangla words and their equivalent terms in English were aligned, and the system achieved an F1-Score of 81.5\%. In a survey paper, Antony \cite{antony2013machine} reports MT approaches and available resources for Indian languages.
A phrase-based SMT has been studied by Islam et al. \cite{islam2010english}, which reports an overall BLEU score of 11.7 and 23.3 for short sentences on the Bangla-English language pair using the EMILLIE-CILL parallel corpora.

In \cite{haffari2009active}, the authors proposed an active learning-based SMT system on the Bangla-English language pair and reported a BLEU score of $\sim$5.7 on the Hansards and EMILLIE corpora. In \cite{pal2011handling}, the authors report a phrased-based SMT approach to handle multiword expressions on the EILMT system and report a BLEU score of 15.12. In another study \cite{ashrafi2013english}, the authors proposed a CFG based approach for assertive sentences to generate rules and translate to the equivalent target rules. 

Following previous work, phrase-based SMT approaches have gained attention. Post et al. in \cite{post2012constructing} report a phrase-based SMT approach on Six Indian Parallel Corpora (SIPC), and with their approach, they report a BLEU score of 12.74 on the Bangla-English language pair. Mishra et al. \cite{mishra2014shata} report another phrase-based SMT system, called Shata-Anuvadak, and evaluated their system on their parallel corpus. Their system achieved a BLEU score of 18.20 for the Bangla-English language pair.
In \cite{khan2017machine}, the authors report a phrase-based SMT technique on the EMILLIE corpora for Indian languages and achieved a BLEU score of 11.8 for the same language pair. For the English–Bangla pair, 
Dandapat and Lewis compared vanilla phrase-based SMT and NMT systems and reported BLEU scores of 16.56 and 20.23, respectively. \cite{dandapat2018training}.
Many-to-one phrase-based SMT has been studied by Banerjee et al. \cite{banerjee2018multilingual}, and with their approach, they report a BLEU score of 13.98 on ILMPC corpora. In \cite{liu2018augmenting}, the authors report a sequence to sequence attention mechanism for Bangla-English MT and achieved a BLEU score of 10.92. In \cite{mumin2019}, the authors proposed a log-linear phrase-based SMT solution named 'shu-torjoma' on the SUPara corpus and reported a 17.43 BLEU score. 


In one of the latest studies, Hasan et al. \cite{hasan2019icbslp} reported both phrase-based SMT and NMT approaches on various parallel corpora. In the said study, authors achieved BLEU scores of 14.82 and 15.62 on SMT, and NMT approaches, respectively, on the ILMPC test set; and a 19.76 BLEU score using the NMT approach with pretrained embedding on the SUPara corpus.
In follow-up work, Hasan et al. \cite{hasan2019neural} also explore NMT for the Bangla-English pair. Authors achieved BLEU scores of 16.58 using a Transformer on the ILMPC test set and a BLEU score of 19.98 using a BiLSTM on the SUPara test set.
NMT has also been studied by Mumin et al. \cite{mumin2019neural} which reports a BLEU score of 22.68, and Hasan et al. \cite{hasan2020not} which reports a BLEU score of 32.10 using a Transformer on the SUPara test set.

For a concise overview, we have provided a list of relevant research on machine translation for Bangla-English MT in Table \ref{tab:mt-table}, which shows that BLEU scores mainly vary within 32.10.

\subsection{Sentiment Classification}
\label{ssec:sentiment_related_work}

\begin{table}[t]
\centering
\caption{Relevant work in the literature for \textbf{sentiment classification}. VAE: Variational Auto Encoder, *C represents number of class labels. Reported results are in Accuracy (Acc). For a few work other metrics has been used, which are mentioned. P: Precision.}
\label{tab:sentiment-table}
\setlength{\tabcolsep}{4pt}
\scalebox{0.85}{
\begin{tabularx}{\textwidth}{@{}XXXX@{}}
\toprule
\multicolumn{1}{c}{\textbf{Paper}} &
\multicolumn{1}{c}{\textbf{Technique}} &
\multicolumn{1}{c}{\textbf{Datasets}} &
\multicolumn{1}{c}{\textbf{Results (Acc)}} \\ \hline
Das et al. \cite{das2010sentiwordnet} & SentiWordNet & 2,234 sentences & 47.6\% \\
Das et al. \cite{das2010phrase} & SVM & 447 sentences & P 70.0\% \\
Chowdhury et al. \cite{6850712} & Unigram & Twitter posts & 93.0\% \\
Taher et al. \cite{taher2018n} & Linear SVM & 9,500 comments & 91.7\% \\
Sumit et al. \cite{sumit2018exploring} & Single layer LSTM & 1.89M sentences & 83.9\% \\
Tripto et al. \cite{tripto2018detecting} & LSTM and CNN & Youtube comments & 65.97\% (3C) and 54.2\% (5C) \\
Vinayakumar \cite{se2015amrita} & Naive Bayes & SAIL & 33.6\% \\
Kumar et al. \cite{kumar2015iit} & SVM & SAIL & 42.2\% \\
Kumar et al. \cite{kumar2020dynamic} & Dynamic model-based features with a random mapping approach & SAIL & 95.4\% \\
Chowdhury et al. \cite{chowdhury2019analyzing} & SVM and LSTM & Movie Reviews & 88.9\% (SVM), 82.4\% (LSTM) \\
Ashik et al. \cite{ashik2019data} & LSTM & Bengali News comments & 79.3\\
Wahid et al. \cite{wahid2019cricket} & LSTM & Cricket comments & 95.0\% \\
Palash et al. \cite{palash2019sentimental} & VAE & Newspaper comments &  53.2\% \\
\bottomrule
\end{tabularx}%
}
\end{table}

 The current state-of-the-art research for Bangla regarding the sentiment classification task includes resource development and addressing the model development challenges. Earlier work includes rule-based and classical machine learning approaches. In \cite{das2010sentiwordnet}, the authors proposed a computational technique of generating an equivalent SentiWordNet (Bangla) from publicly available English sentiment lexicons and an English-Bangla bilingual dictionary with very few easily adaptable noise reduction techniques. The classical algorithms used in different studies include Bernoulli Naive Bayes (BNB), Decision Tree, SVM, Maximum Entropy (ME), and Multinomial Naive Bayes (MNB) \cite{9084046,banik2018evaluation,chowdhury2019analyzing}. 
In \cite{islam2016supervised}, the authors developed a polarity detection system on textual movie reviews in Bangla by using two widely used machine learning algorithms: NB and SVM, and providing comparative results. In another study, the authors used NB with rules for detecting sentiment in Bengali Facebook statuses~\cite{islam2016supervised}. In \cite{6850712}, the authors developed a dataset using semi-supervised approaches and designed models using SVM, and Maximum Entropy \cite{6850712}. 

The work related to the use of deep learning algorithms for sentiment analysis include \cite{hassan2016sentiment,sharfuddin2018deep,tripto2018detecting,ashik2019data,DBLP:journals/corr/abs-2004-07807}. In \cite{tripto2018detecting}, the authors used LSTMs and CNNs with an embedding layer for both sentiment and emotion identification from YouTube comments. The study in \cite{ashik2019data} provides a comparative analysis using both classical -- SVM, and deep learning algorithms -- LSTM and CNN, for sentiment classification of Bangla news comments.
The study in \cite{DBLP:journals/corr/abs-2004-07807} integrated word embeddings into a Multichannel Convolutional-LSTM (MConv-LSTM) network for predicting different types of hate speech, document classification, and sentiment analysis for Bangla. Due to the availability of romanized Bangla texts in social media, the studies in \cite{hassan2016sentiment,sharfuddin2018deep} use LSTM to design and evaluate the model for sentiment analysis. In \cite{alam2017sentiment}, authors used a CNN for sentiment classification of Bangla comments.
The studies in \cite{rahman2018datasets} and \cite{mahtab2018sentiment} analyze user sentiment on Cricket comments from online news forums.

For sentiment analysis, there has been significant work in terms of resource and model development. In Table \ref{tab:sentiment-table}, we report a concrete summary highlighting techniques, datasets, and the reported results in different studies. 

\subsection{Emotion Classification}
\label{ssec:emotion_related_work}

The work in emotion classification is relatively sparse compared to sentiment classification for Bangla content. To this effect, Das et al. \cite{das2010sentiwordnet} developed WordNet affect lists for Bangla, which is adapted from English affect word-lists. 
Tripto et al. \cite{tripto2018detecting} used LSTM and CNN with an embedding layer for emotion identification from YouTube comments. Their proposed approach shows an accuracy of 59.2\% for emotion classification.

\subsection{Authorship Attribution}
\label{ssec:authorship_related_work}

Authorship attribution is another interesting research problem in which the task is to identify original authors from the text. The research work in this area is comparatively low. In \cite{khatun2019authorship}, the authors developed a dataset and experiment with character level embedding for authorship attribution. Using the same dataset, Alam et al. \cite{alam2020bangla} fine-tune multi-lingual transformer models for the authorship identification task and report an accuracy of 93.8\%.

\subsection{News Categorization}
\label{ssec:news_related_work}
The News Categorization task is one of the earliest pieces of work in NLP in a number of languages. However, compared to other languages, not much has been done for Bangla. One of the earliest studies for Bangla news categorization is by Mansur et al. \cite{Mansur2010}, which looked at character-level n-gram based approaches. They reported different n-grams results in terms of frequency, normalized frequency, and ranked frequency. Their study showed that when n is increased from 1 to 3, the performance increases. However, from a value of 3 to 4 or more, the performance decreases.

The study of Mandal et al. \cite{mandal2014supervised} used four supervised learning methods: Decision Tree(DT), K-Nearest Neighbour (KNN), Naïve Bayes (NB), and Support Vector Machine (SVM) for the categorization of Bangla news from various Bangla websites. Their approach includes tokenization, digit removal, punctuation removal, stop words removal, and stemming, followed by feature extraction using normalized tf-idf weighting and length normalization. They reported precision, recall and F-score for every category. They also reported an average (macro) F-score for all of the four machine learning algorithms: DT(80.7), KNN(74.2), NB(85.2), and SVM(89.1). In \cite{8554811}, the authors extracted tf-idf features and trained the classifier using Random Forest, SVM with linear and radial basis kernel, K-Nearest Neighbor, Gaussian Naive Bayes, and Logistic Regression. They have created a large Bangla text dataset and made it publicly available. 
Another study that is similar has been done by Alam et al. \cite{8554382} using a corpus of size $\sim$3,76,226 of Bangla news articles.
The study conducted experiments using Logistic Regression, Neural Network, NB, Random Forest, and Adaboost by utilizing textual features such as word2vec, tf-idf (3000 word vector), and tf-idf (300 word vector). They obtained the best results, an F1 of 0.96, using word2vec representation and neural networks. 

%% file: sections/methodology.tex
\section{Methods}
\label{sec:methods}

We use different multilingual and monolingual transformer-based language models in our experiments. For monolingual models, we make use of Bangla language models trained in Indic-Transformers \cite{jain2020indic}. Indic-Transformers consist of language models trained for three Indian languages: Hindi, Bangla, and Telugu. The authors introduced four variants of the monolingual language model: BERT, DistilBERT, RoBERTa, and XLM-RoBERTa.

In this section, we briefly describe different language models we have used and task-specific modifications that were done for fine-tuning them. 

\subsection{Pretrained Language Models}
\subsubsection{BERT} 
BERT \cite{devlin2018bert} is designed to learn contextualized word representation from unlabeled texts by jointly conditioning on the left and right contexts of a token. It uses the encoder part of the transformer architecture introduced in \cite{NIPS2017_7181}. Two objective functions are used during the pretraining step: 
\begin{itemize}
    \item Masked language model (MLM): Some fraction of the input tokens are randomly masked, and the objective is to predict the vocabulary ID of the original token in that position. The bidirectional nature ensures that the model can effectively use both past and future contexts for this task.
    \item Next sentence prediction (NSP): This is a binary classification task where given two sentences, the goal is to decide whether the second sentence immediately follows the first sentence in the original text. Positive sentences are created by taking consecutive sentences from the text, and negative sentences are created by taking sentences from two different documents. 
\end{itemize}




The multilingual variant of BERT (mBERT) is trained using the Wikipedia corpus of the most extensive languages. Data is sampled using an exponentially smoothed weighting to address differences among the corpus size of different languages, ensuring that high resource languages like English are under-sampled compared to low resource languages. Word counts are weighted similarly so that words from low-resource languages are represented adequately in terms of vocabulary.

Two commonly used variants of BERT models are BERT-base and BERT-large. BERT-base model consists of 12 layers, 768 hidden dimensions, and 12- self-attention heads. BERT-large variant has 24 layers, 1,024 hidden dimensions, and 16 self-attention heads.

We use three different BERT models in our experiments:
\begin{enumerate}
    \item BERT-m: We use the pretrained multilingual BERT model available in HuggingFace's transformer library.\footnote{\url{https://huggingface.co/bert-base-multilingual-cased}} This model is trained using the top 104 languages with the largest Wikipedia entries, including Bangla. 
    \item BERT-bn: This is a monolingual Bangla BERT model trained using the same architecture as the BERT-base model.\footnote{\url{https://huggingface.co/sagorsarker/bangla-bert-base}} Bangla common crawl corpus, and Wikipedia dump dataset are used to train this language model. It has 102025 vocabulary entries.
    \item Indic-BERT: This is the monolingual BERT model from Indic-Transformers trained using $\sim$3 GB training data.\footnote{\url{https://huggingface.co/neuralspace-reverie/indic-transformers-bn-bert}}
\end{enumerate}

\begin{table}[]
\caption{Configurations for different Transformer models used in the experiments. 
}
\label{tab:model-stats}
\setlength{\tabcolsep}{4pt}
\scalebox{0.7}{
\begin{tabular}{@{}lccccccc@{}}
\toprule
\textbf{Model Name} & \textbf{Model Type} & \textbf{\begin{tabular}[c]{@{}c@{}}\#Parameters \\ (Millions)\end{tabular}} & \textbf{\begin{tabular}[c]{@{}c@{}}Mono/Multi \\ lingual\end{tabular}} & \textbf{\begin{tabular}[c]{@{}c@{}}Vocab \\ size\end{tabular}} & \textbf{\begin{tabular}[c]{@{}c@{}}Hidden \\ Size\end{tabular}} & \textbf{\begin{tabular}[c]{@{}c@{}}\#Hidden \\ layers\end{tabular}} & \textbf{\begin{tabular}[c]{@{}c@{}}\#Attention \\ heads\end{tabular}} \\ \midrule
\textbf{Bangla Electra} & base & \,13.4 & mono & 29,898 & 256 & 12 & 4 \\
\textbf{Indic-BERT} & base & 134.5 & mono & 100,000 & 768 & 12 & 12 \\
\textbf{Indic-DistilBERT} & base & 66.4 & mono & 30,522 & 768 & 6 & 12 \\
\textbf{Indic-RoBERTa} & $**$ & 83.5 & mono & 52,000 & 768 & 6 & 12 \\
\textbf{Indic-XLM-RoBERTa} & $**$ & 134.5 & mono & 100,002 & 768 & 8 & 12 \\
\textbf{BERT-bn} & base & 164.4 & mono & 102,025 & 768 & 12 & 12 \\
\textbf{BERT-m} & base & 177.9 & multi & 119,547 & 768 & 12 & 12 \\
\textbf{DistilBERT-m} & base & 134.7 & multi & 119,547 & 768 & 6 & 12 \\
\textbf{XLM-RoBERTa} & large & 559.9 & multi & 250,002 & 1,024 & 24 & 16 \\
\textbf{Transformer\tablefootnote{Only used for MT task}} & base & 253.1 & multi & 256,360(BN), 151,752(EN) & 512 & 6 & 8 \\
 \bottomrule
\end{tabular}
}
\end{table}

\subsubsection{RoBERTa} 
RoBERTa \cite{xlm-roberta} improves upon BERT by proposing several novel training strategies, including 
\begin{enumerate*}
\item training the model longer with more data
\item using a larger batch size
\item removing the next sentence prediction task and only using MLM loss
\item training on longer sequences
\item generating the masking pattern dynamically.
\end{enumerate*}
These modifications allow RoBERTa to outperform BERT on different downstream language understanding tasks consistently. 

XLM-RoBERTa \cite{xlm-roberta} is the multilingual counterpart of RoBERTa trained with a multilingual MLM. It is trained in one hundred languages using 2.5 terabytes of filtered Common Crawl data. Like RoBERTa, it provides substantial gain over the multilingual BERT model, especially on low resource languages.

We used three different RoBERTa models in our experiments:

\begin{enumerate}
    \item XLM-RoBERTa: We use the XLM-RoBERTa large model from HuggingFace's transformer library.\footnote{\url{https://huggingface.co/xlm-roberta-large }}
    
    \item Indic-RoBERTa: This is the monolingual RoBERTa language model trained on $\sim$6 GB training corpus from Indic-Transformers.\footnote{\url{https://huggingface.co/neuralspace-reverie/indic-transformers-bn-roberta}} 
    
    \item Indic-XLM-RoBERTa: This XLMRoBERTa model is pre-trained on $\sim$3 GB of monolingual training corpus.\footnote{\url{https://huggingface.co/neuralspace-reverie/indic-transformers-bn-xlmroberta}} 
\end{enumerate}



\subsubsection{DistilBERT}
DistilBERT \cite{sanh2019distilbert} is trained using knowledge distillation from BERT. This model is 40\% smaller and 60\% faster while retaining 97\% of the language understanding capabilities of the BERT model. The training objective used is a linear combination of distillation loss, supervised MLM loss, and cosine embedding loss. 

We use two variants of the DistilBERT model in our experiments:

\begin{enumerate}
    \item DistilBERT-m: This model is distilled from the multilingual BERT model.\footnote{\url{https://huggingface.co/distilbert-base-multilingual-cased}} Similar to BERT-m, it is trained on 104 languages from Wikipedia. 
    \item Indic-DistilBERT: This DistilBERT model is trained on $\sim$6 GB of monolingual training corpus.\footnote{\url{https://huggingface.co/neuralspace-reverie/indic-transformers-bn-distilbert}}
\end{enumerate}

\subsubsection{Electra}
Electra \cite{clark2020electra} is trained using a sample-efficient pre-training task called replaced token detection. In this approach, instead of masking input tokens, they are replaced with alternatives sampled from a generator network. Then a discriminator model is trained to predict whether a generator sample replaced each token or not.  This approach allows the model to learn better representation while being compute-efficient. 
We use a monolingual Bangla Electra model trained on a 5.8 GB web crawl corpus and 414 MB Bangla Wikipedia dump.\footnote{\url{https://huggingface.co/monsoon-nlp/bangla-electra}}

In Table \ref{tab:model-stats} we present the specific configurations of the models we used in our study. For some pre-trained models, authors have not used original architectures as highlighted with $**$ in \textit{Model Type} column. The table shows that the vocab size for multilingual models is more extensive than monolingual models as they contain words from different languages. Among the models, Electra is the smallest in terms of vocab size, hidden units, number of layers, and number of attention heads, whereas XLM-RoBERTa is the largest. 

\subsection{Task-specific Fine-Tuning}
We take hidden layer embeddings from the pretrained models and add additional layers for the specific task at hand. We then fine-tune the entire network using the dataset in an end-to-end manner. Even though we experiment with nine different tasks, they can be divided into three broad categories from the modeling perspective.

\subsubsection{Text Classification} We consider \textit{sentiment}, emotion, \textit{authorship attribution}, and news categorization as a text classification problem, where text can be a document, article or social media post. For these tasks, we use sentence embeddings (usually the start of sequence token in the transformers) and use it for classification. We add a linear layer to predict the output class distribution for the task. The linear layer is preceded by an additional hidden layer for RoBERTa models.

\subsubsection{Token/Sequence Classification} We consider PoS tagging, lemmatization, NER, and punctuation restoration as token classification tasks. We use the hidden layer embeddings obtained from the transformers as input for a bidirectional LSTM layer. The outputs from the forward and backward LSTM layers are concatenated at each time-step and fed to a fully connected layer to predict the token distribution, allowing the network to use both past and future contexts for prediction effectively. 

\subsubsection{Machine Translation} 
For the machine translation task, we use a transformer base model to train the data and byte pair encoding for handling rare words. Each sentence starts with a special \textit{start of sentence token} and ends with a \textit{end of sentence token}. For the MT task, these are respectively \textbf{[START]} and \textbf{[END]} tokens. Each encoder layer consists of a multi-head attention layer followed by a fully connected feed forward layer, in which the decoder layer consists of a masked multi-head attention and an encoder attention layer followed by a fully connected feed forward layer. 





\subsection{Evaluation}
\label{ssec:evaluation}
 We computed the weighted average precision (P), recall (R) and F1-measure (F1) to measure the performance of each classifier. We chose a weighted metric, which takes care of the class imbalance problem. For the MT task, we computed the Bilingual Evaluation Understudy (BLEU) score \cite{papineni2002bleu} to evaluate the performance of automatic translations.

%% file: sections/experiments.tex
\section{Experiments}
\label{sec:experiments}

\input{sections/experiments/pos}
\input{sections/experiments/lemma}

\input{sections/experiments/ner}

\input{sections/experiments/punc_restore}

\input{sections/experiments/mt}

\input{sections/experiments/affective_behavior}

\input{sections/experiments/news_categorization}

%% file: sections/experiments/pos.tex
\subsection{Parts of Speech}
\label{ssec:exp_pos}

\subsubsection{Dataset}
\label{sssec:dataset_pos}
For the POS task, we used the following three datasets for training and evaluating the models. In the sections below, we provide brief details for each dataset. 

\begin{enumerate}
\item \textbf{LDC Corpus~\cite{baskaran2008common,baliLDC2010T16}:} The LDC corpus is publicly available through LDC. It has been developed by Microsoft Research (MSR), India, for linguistic research. It consists of three-level annotations i.e. lexical category, type, and morphological attribute. For the current study, we only utilized POS tags comprising 30 tags. The entire corpus consists of 7,393 sentences corresponding to 102,937 tokens. The text in the corpus was collected from blogs, Wikipedia articles, and other sources in order to have variation in the text. More details of the said tag set can be found in the annotation guideline included with the corpus, and also found in \cite{baskaran2008common}. 

\item \textbf{IITKGP POS Tagged Corpus~\cite{iitgpgpostagging}:} The IITKGP POS Tagged corpus consists of a tagset comprising 38 tags, developed by Microsoft Research in collaboration with IIT Kharagpur and several institutions in India \cite{iitgpgpostagging}. For the current study we mapped this tagset with the tagset of LDC corpus to make it consistent. The dataset consists of 5,473 sentences and 72,400 tokens.

\item \textbf{CRBLP POS Tagged Corpus~\cite{ummi2008developing}:} The CRBLP POS Tagged corpus consists of $\sim20K$ tokens, from 1176 sentences, manually tagged based on the tagset proposed in \cite{hayder2007research,mahmud2009syntactic}. The articles were collected from BDNews24\footnote{\url{www.bdnews24.com}}, one of the most widely circulated newspapers in Bangladesh. For training and evaluation, we also mapped the tagset to align with the other two datasets mentioned above.  
\end{enumerate}

\begin{figure*}[ht]
\centering
\includegraphics[width=5.0in]{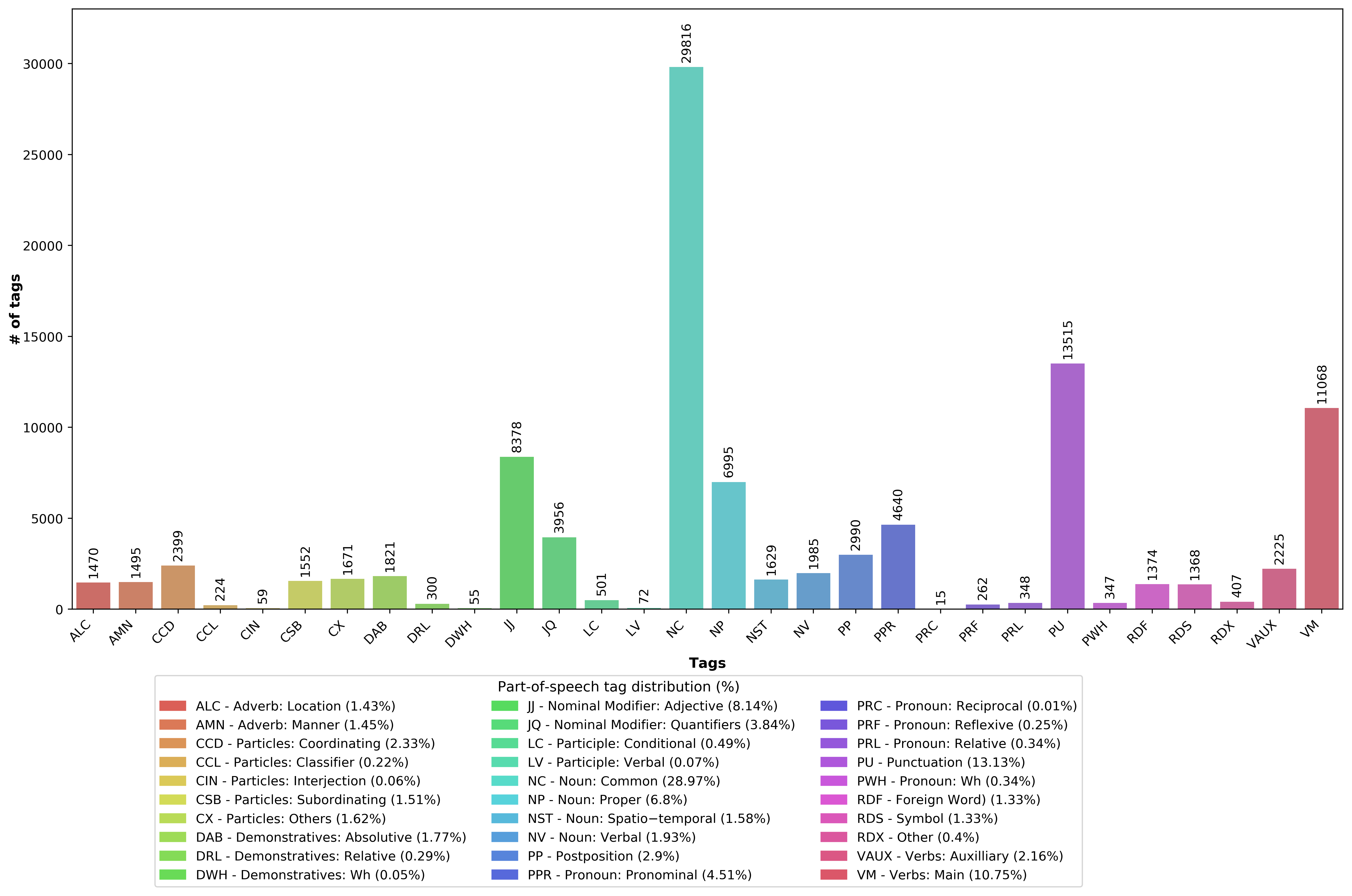}
\caption{POS tag distribution in LDC corpus.}
\label{fig:ldc-pos-tag-dist}
\end{figure*}

\begin{figure*}[ht]
\centering
\includegraphics[width=5.0in]{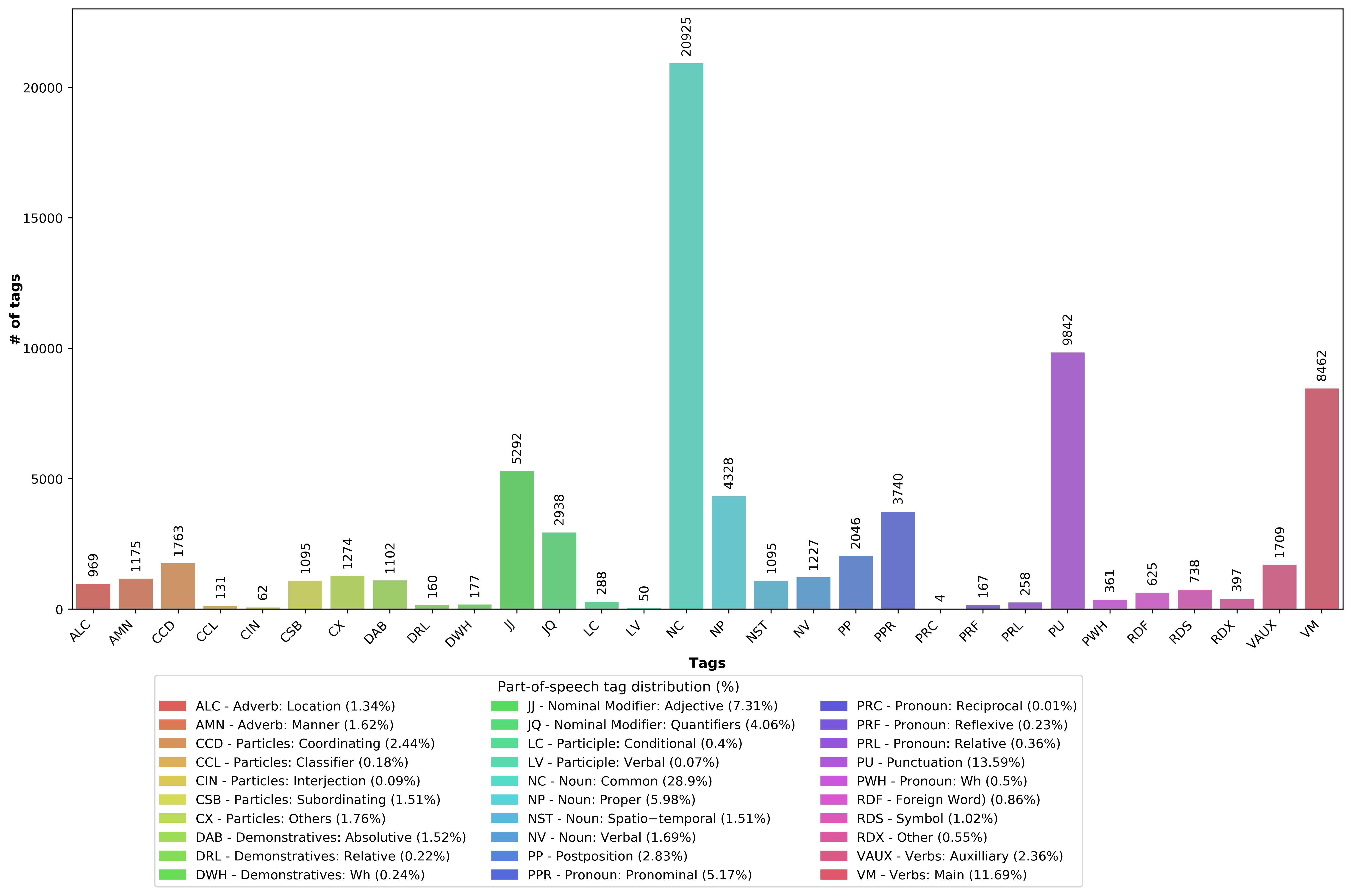}
\caption{POS tag distribution in IITKGP corpus.}
\label{fig:iitkpg-pos-tag-dist}
\end{figure*}

\begin{figure*}[ht]
\centering
\includegraphics[width=5.0in]{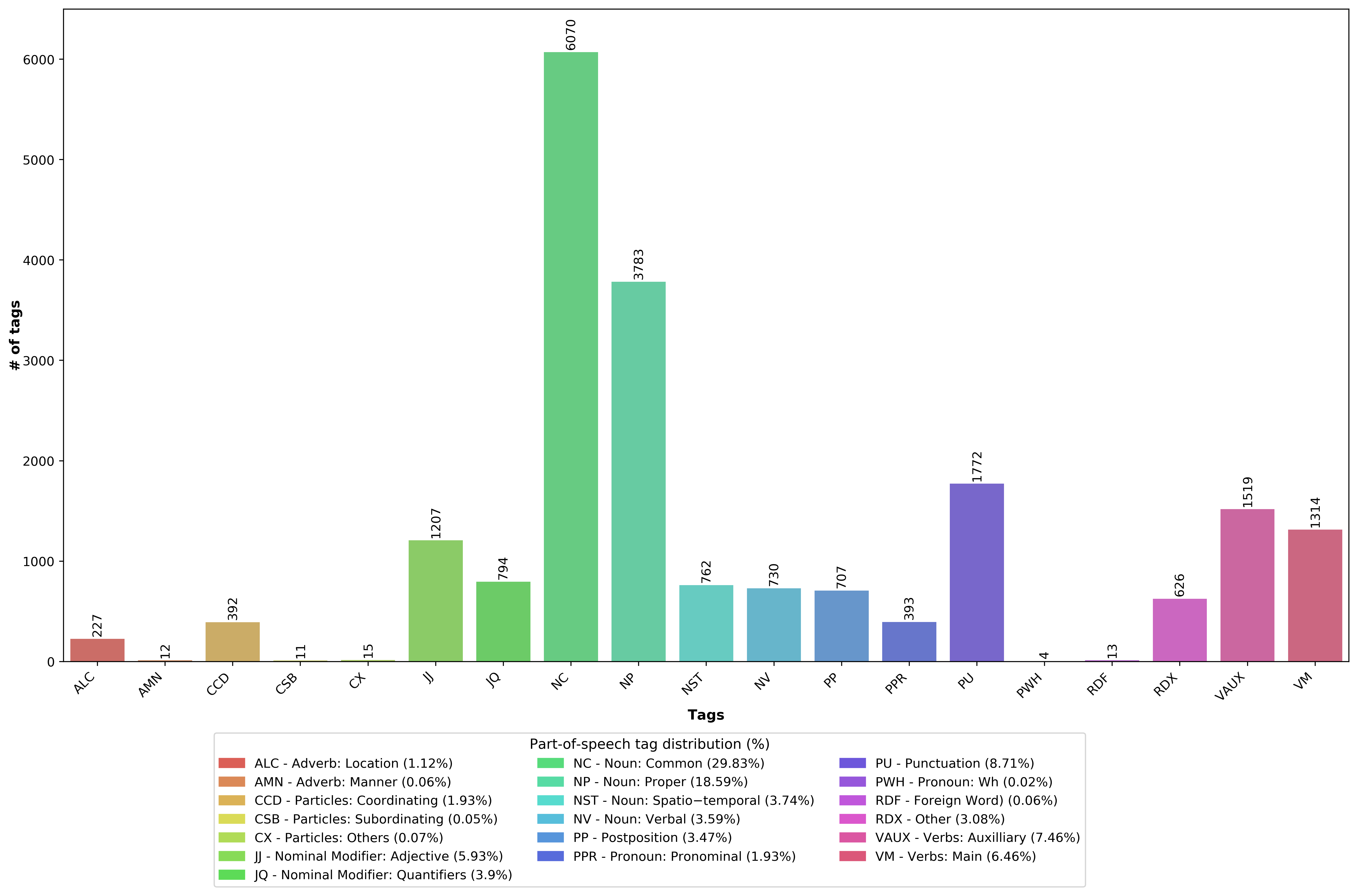}
\caption{POS tag distribution in CRBLP corpus.}
\label{fig:bangla-rabia-pos-tag-dist}
\end{figure*}

\begin{table}[h]
\centering
\caption{Training, development and test data split for \textbf{POS tagging} task. ``Bangla 1" and ``Bangla 2" are two sets in the original LDC distribution.}
\label{table:data-split-pos-tagging}
\setlength{\tabcolsep}{4pt}
\scalebox{0.8}{
\begin{tabular}{lrrl}
\toprule
\multicolumn{1}{c}{\textbf{Data set}} & \multicolumn{1}{c}{\textbf{\# of Sent}} & \multicolumn{1}{c}{\textbf{\# of Token (\%)}} & \multicolumn{1}{c}{\begin{tabular}[l]{@{}l@{}}\textbf{Set from Original Source}\end{tabular}} \\ \midrule
\textbf{Train} & 4,575 & 62,048 ($\sim$60\%) & \begin{tabular}[l]{@{}l@{}}``Bangla 1" data set +\\ (1 to 4) from ``Bangla 2" data set\end{tabular} \\ 
\textbf{Dev} & 1,455 & 20,435 ($\sim$20\%) & (5 to 10) from ``Bangla 2" data set \\ 
\textbf{Test} & 1,368 & 20,437 ($\sim$20\%) & (11 to 17) from ``Bangla 2" data set \\ \bottomrule
\end{tabular}
}
\end{table}  

In Figure \ref{fig:ldc-pos-tag-dist}, we present the POS tag distribution, in percentage, for the whole LDC corpus. The distribution is very low for some tags, for example, CIN (Particles: Interjection) - 59, DWH (Demonstratives: Wh) - 55, LV (Participle: Verbal) - 72, and PRC (Pronoun: Reciprocal) - 15. Such a skewed tag distribution also affects the performance of the automatic sequence labeling (tagging) task. In Figure \ref{fig:iitkpg-pos-tag-dist} and \ref{fig:bangla-rabia-pos-tag-dist} we present the POS tag distribution for the IITKGP, and the CRBLP POS Tagged Corpus, respectively. Across all datasets, the distribution of noun, main verb, and punctuation are higher. 

\subsubsection{Training}
\label{sssec:traning_pos}
For training, fine-tuning, and evaluating the models, we used the same LDC Corpus data splits (i.e., training, development, and test set) reported in \cite{alam2016bidirectional}, also shown in Table \ref{table:data-split-pos-tagging}. In the original distribution, the data set appears in two sets ``Bangla 1" and ``Bangla 2". It has been divided by maintaining the file numbers and the associated set where the distribution of the data split is $60\%$, $20\%$ and $20\%$ of the tokens, for the training, development, and test set, respectively. There are many unknown words in the LDC Corpus data split, i.e., words that are not present in the training set. About $\sim51\%$ tokens in the development and test sets are of unknown type.

As additional experiments, we combined {\em(i)} LDC training set, {\em(ii)} IITKGP POS Tagged Corpus, and {\em(iii)} CRBLP POS Tagged Corpus as consolidated training sets. We used the LDC development set for fine-tuning the models and evaluated it using the LDC test set for all experiments. We fine-tuned the pre-trained models as discussed in section \ref{sec:methods}.

%% file: sections/experiments/lemma.tex
\subsection{Lemmatization}
\label{ssec:exp_lemmatization}


\subsubsection{Dataset}
\label{sssec:dataset_lemma}
We used the corpus reported in \cite{chakrabarty2017context}. The raw text was collected from a collection of Rabindranath Tagore’s short stories and news articles from various domains. The authors annotated the raw text to prepare a gold lemma dataset.\footnote{\url{https://www.isical.ac.in/~utpal/resources.php}} In table \ref{table:lemma_data_split}, we present the data split of our experiment, in which we used 70\%, 15\% and 15\% sentences for training, development and test set, respectively.

\begin{table}[h]
\centering
\caption{Training, development and test data split for \textbf{Lemmatization} task.}
\label{table:lemma_data_split}
\begin{tabular}{lrr}
\toprule
\multicolumn{1}{c}{\textbf{Data set}} & \multicolumn{1}{c}{\textbf{\# of Sent}} & \multicolumn{1}{c}{\textbf{\# of Token}} \\\midrule
\textbf{Train} & 1,191 ($\sim$70\%) & 14,091 \\ 
\textbf{Dev} & 256 ($\sim$15\%) & 3,028  \\ 
\textbf{Test} & 255 ($\sim$15\%) & 3,135 \\ \midrule
\textbf{Total} & 1,702 & 20,254 \\ \bottomrule
\end{tabular}
\end{table}

\subsubsection{Training}
\label{sssec:dataset_lemma}
For training, we used all the transformer models discussed in Section \ref{sec:methods}. We also used the same fine-tuning procedures as other token classification tasks (e.g., POS).

%% file: sections/experiments/ner.tex
\subsection{Named Entity Recognition}
\label{ssec:exp_ner}

\subsubsection{Dataset}
\label{sssec:dataset_ner}
We used the corpus reported in \cite{chowdhury2018towards}, referred to as the Bangla Content Annotation Bank (B-CAB).\footnote{\url{https://github.com/Bangla-Language-Processing/Bangla-Content-Annotation-Bank}} The text for the corpus has been collected from various popular newspapers in Bangladesh (e.g., Prothom-Alo\footnote{\url{https://www.prothomalo.com/}}). It consists of $35$ news articles, $\approx 35 K$ words, $2,137$ sentences with a vocabulary size of $|V|\approx 10 K$. The topics range from politics, sports, entertainment etc. The annotated dataset consists of the following seven entity types:
\begin{itemize}
\item \textbf{Person (PER):} Person entities are only defined for humans. A person entity can be a single individual or a group. 
\item \textbf{Location (LOC):} Location entities are defined as geographical entities, which include geographical areas and landmasses, bodies of water, and geological formations.
\item \textbf{Organization (ORG):} Organization entities are defined by corporations, agencies, and other groups of people. 
\item \textbf{Facility (FAC):} Facility entities are defined as buildings and other permanent human-made structures. 
\item \textbf{Time (TIME):} Time entities represent absolute dates and times. It includes duration, days of the week, month, year, and time of day. 
\item \textbf{Units (UNITS):} Units are mentions that include money, number, rate, and age. 
\item \textbf{Misc (MISC):} Misc entities are any entities that do not fit into the above entities.
\end{itemize}

\begin{figure}[h]
\centering
\includegraphics[width=4.2in]{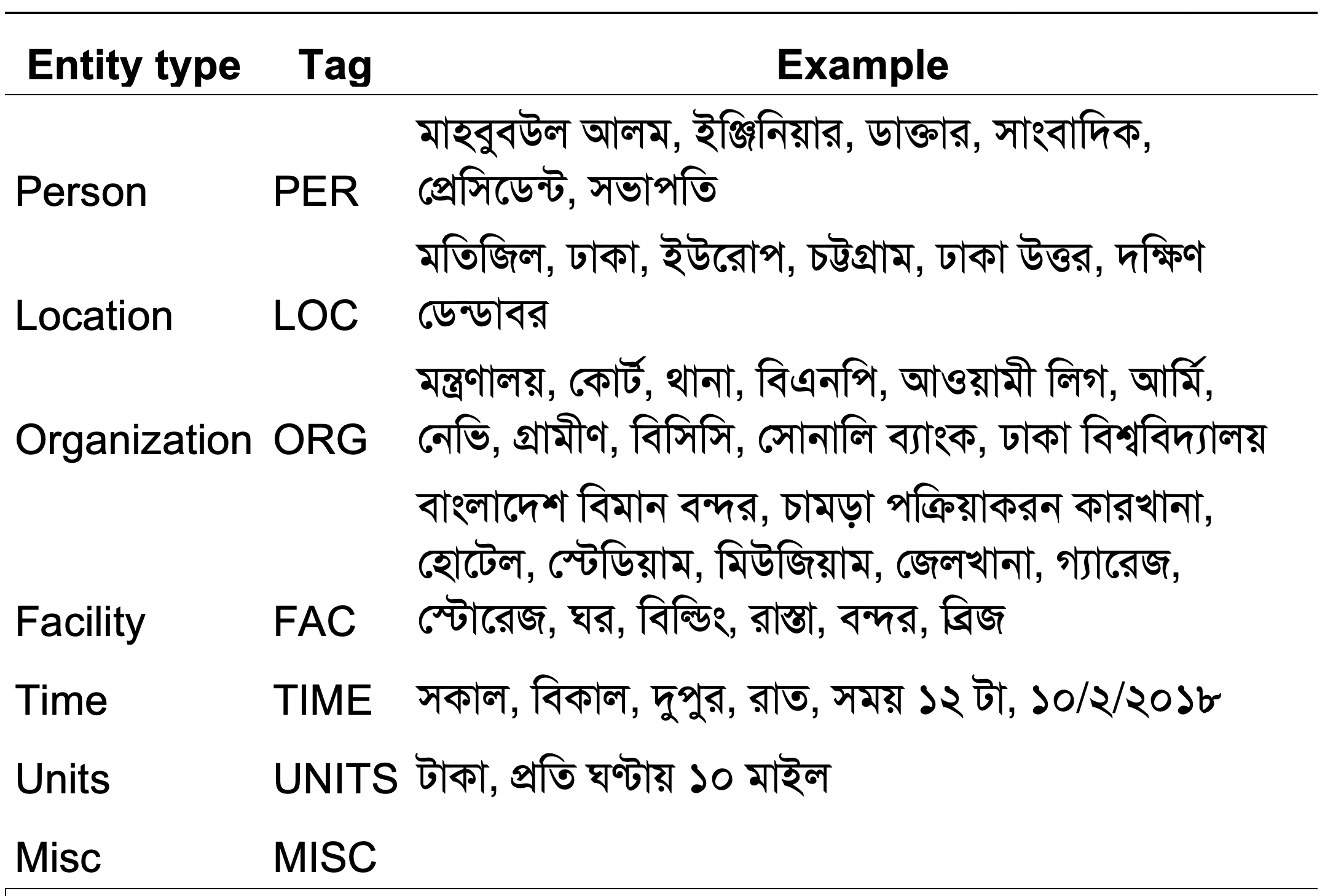}
\caption{Example of entity type and tags.}
\label{fig:entity-type-tag-dist}
\end{figure}



\begin{figure}[h]
\centering
\includegraphics[width=4.2in]{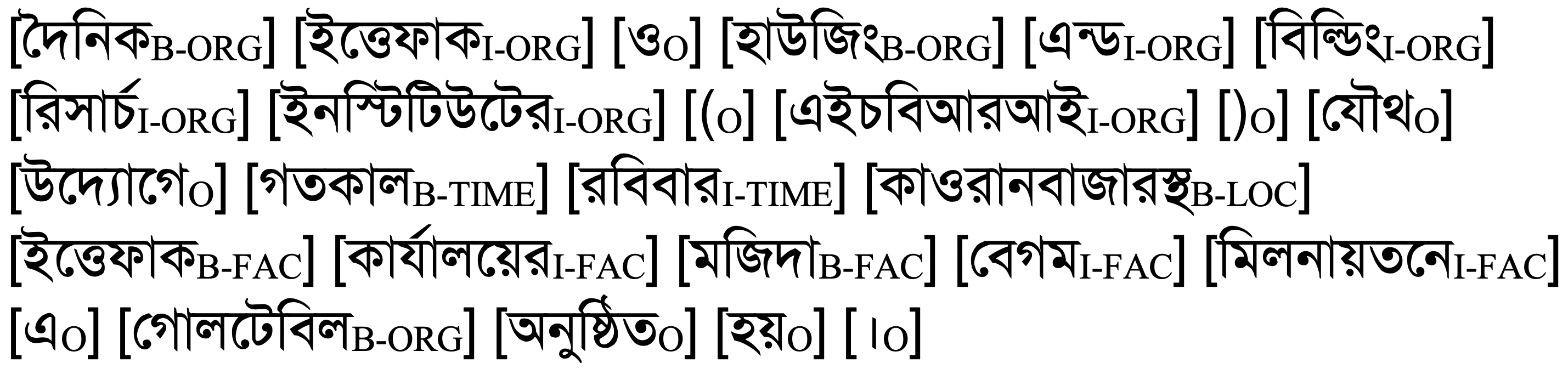}
\caption{Example of an annotated sentence.}
\label{fig:ner_annotation_example}
\end{figure}

\begin{figure}[h]
\centering
\includegraphics[width=4.0in]{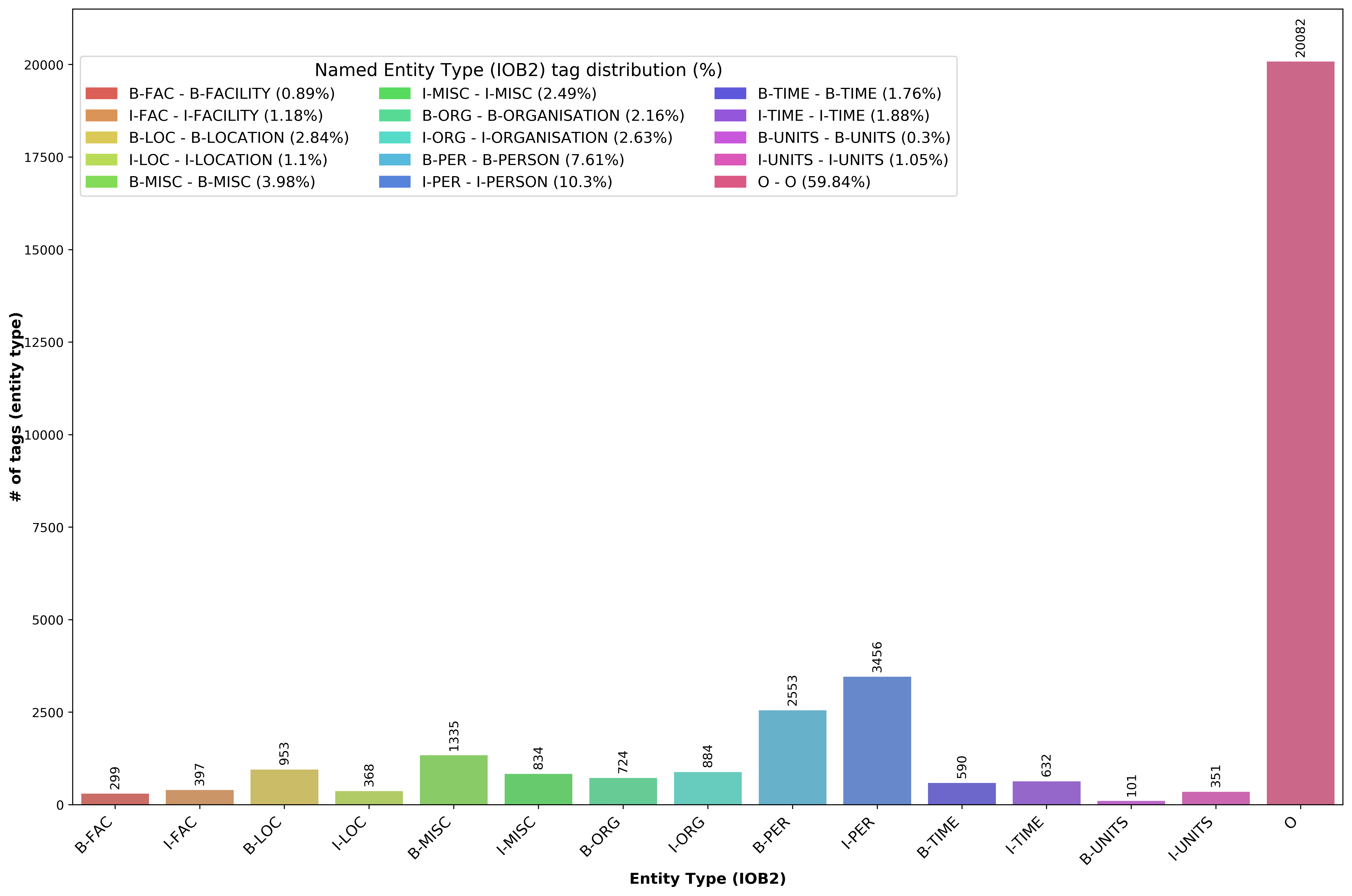}
\caption{Entity type with IOB2 tag distribution.}
\label{fig:iob-tag-dist}
\end{figure}

In Figure \ref{fig:entity-type-tag-dist}, we report examples of each entity type and tag. The annotation has been prepared in IOB2 format as shown in Figure \ref{fig:ner_annotation_example}. We present the distribution of the entity type in IOB2 format in Figure \ref{fig:iob-tag-dist}, which shows that more than $50\%$ of the textual content are non-entity mentions tagged as $O$, which is a typical scenario for any named entity corpus. Among the entity types, \textit{person} type entities are higher. From the figure, we observe that entity type distribution across datasets is representative of the machine learning experiments. In Table \ref{table:token_level_entity_mention_stat}, we provide token level statistics for each entity type. On average, two to three tokens per entity mention for each entity type. We also observed that in some cases, the number of tokens went up to ten to fifteen due to the fact that the title and the subtitle are associated with person entity mentions. Such entity mentions pose various challenges for an automated recognition system. 


\begin{table}[h]
\centering
\caption{Statistics with the number of token in \textbf{entity mentions} of each entity type.}
\label{table:token_level_entity_mention_stat}
\begin{tabular}{@{}lrrr@{}}
\toprule
\multicolumn{1}{c}{\textbf{Entity types}} & \multicolumn{1}{c}{\textbf{Avg.}} & \multicolumn{1}{c}{\textbf{Std.}} \\ \midrule
FACILITY & 2.5 & 1.7 \\
LOCATION & 1.6 & 1.3 \\
MISC & 2.0 & 1.4 \\
ORGANISATION & 2.5 & 1.4 \\
PERSON & 2.9 & 1.9 \\
TIME & 2.3 & 1.3 \\
UNITS & 2.5 & 2.1 \\ \bottomrule
\end{tabular}
\end{table}

\subsubsection{Training}
\label{sssec:traning_ner}
In order to train the model, we use the same data splits reported in \cite{chowdhury2018towards}. In Table \ref{table:data_stat_ner}, we provide the statistics of the tokens and sentences for different splits. The data split consists of $\sim70\%$, $\sim10\%$, and $\sim20\%$ of the tokens for the training, development, and test set, respectively. In Table \ref{tab:entity-type-tag-dist}, we present entity type distribution for different splits, which shows overall \textit{UNITS} entity type is under-representative. 

For the experiments, we used the same transformer models discussed in Section \ref{sec:methods}. For this task, we used a maximum sequence length of 256, a batch size of 8, and a learning rate of 1e-5. The models were trained using the Adam optimization algorithm for 10 epochs.

\begin{table}[h]
\centering
\caption{Statistics of the annotated \textbf{NER} dataset for the training, development and test data split. The first row represents the total number of sentences in each set. Row 2-5 represents the total, the average, the standard deviation, and the maximum number of tokens in each set.}
\label{table:data_stat_ner}
\begin{tabular}{@{}lrrr@{}}
\toprule
\textbf{Metric} & \multicolumn{1}{l}{\textbf{Train}} & \multicolumn{1}{l}{\textbf{Dev}} & \multicolumn{1}{l}{\textbf{Test}} \\ \midrule
\# of sentences & 1,510 & 200 & 427 \\
Total & 24,377 & 2,636 & 6,546 \\
Average & 16.14 & 13.18 & 15.33 \\
Std. deviation & 9.73 & 7.83 & 13.45 \\ 
Max & 98 & 53 & 85 \\\bottomrule
\end{tabular}
\end{table}


\begin{table}[h]
\centering
\caption{Training, development and test split for \textbf{NER} (Entity type distribution) task.}
\label{tab:entity-type-tag-dist}
\begin{tabular}{@{}lrrr@{}}
\toprule
\multicolumn{1}{c}{\textbf{Entity type}} & \multicolumn{1}{c}{\textbf{Train}} & \multicolumn{1}{c}{\textbf{Dev}} & \multicolumn{1}{c}{\textbf{Test}} \\ \midrule
LOCATION & 738 & 177 & 177 \\
PERSON & 1,954 & 440 & 440 \\
FACILITY & 190 & 71 & 71 \\
MISC & 960 & 234 & 234 \\
TIME & 446 & 76 & 76 \\
ORGANISATION & 593 & 91 & 91 \\
UNITS & 60 & 33 & 33 \\ \bottomrule
\end{tabular}%
\end{table}

%% file: sections/experiments/punc_restore.tex
\subsection{Punctuation Restoration}
\label{ssec:exp_punc}

\subsubsection{Dataset}
\label{sssec:dataset_punc_restore}
 For this task, we used the dataset reported in \cite{alam-etal-2020-punctuation}\footnote{https://github.com/xashru/punctuation-restoration} for the punctuation restoration task. The dataset consists of train, development, and test splits prepared from a publicly available corpus of Bangla newspaper articles \cite{DBLP:journals/corr/abs-1911-07613}. Additionally, the authors prepared two test datasets from manual and ASR transcribed texts. These were collected from 65 minutes of speech excerpts extracted from four Bangla short stories. There are four labels including three punctuation marks: {\em(i) Comma:} includes commas, colons and dashes, {\em(ii) Period:} includes full stops, exclamation marks and semicolons, {\em(iii) Question:} only question mark, and {\em(iv) O:} for any other token.

In Table \ref{tab:data-bangla-puncres}, we present the distributions of the labels for the dataset. In parenthesis, we provide the percentage of the punctuation. The distribution of \textit{Question} is low (less than 1\%) in the news data but much higher in the Bangla manual and ASR transcriptions. The low distribution can be attributed to the texts being selected from short stories in which people often engage in conversation and ask questions in dialogue. The distribution of \textit{Period} is also higher in the Bangla manual and ASR transcriptions. The higher distribution results in a much smaller average sentence length in these datasets, as shown in Table \ref{tab:data-bangla-puncres}.

\subsubsection{Training}
\label{sssec:exp_settings_punc_restore}
Given that we used the same data splits as reported in \cite{alam-etal-2020-punctuation}, in which experiments have been conducted using multilingual transformer models, hence, for the punctuation restoration task, we only used monolingual models and compared the results. For the experiments, we used a maximum sequence length of 256, a batch size of 8, and a learning rate of 1e-5. The models were trained using the Adam optimization algorithm for 10 epochs. 

\begin{table}
\centering
\caption{\textbf{Punctuation} data distributions along with average sentence length (Avg.) and standard deviation (Std.). The number in parenthesis represents percentage.}
\label{tab:data-bangla-puncres}
\scalebox{0.80}{
\begin{tabular}{lrrrrrrr}
\toprule
\multicolumn{1}{l}{\textbf{Dataset}} & \multicolumn{1}{c}{\textbf{Total}} & \multicolumn{1}{c}{\textbf{Period}} & \multicolumn{1}{c}{\textbf{Comma}} & \multicolumn{1}{c}{\textbf{Question}} & \multicolumn{1}{c}{\textbf{Other (O)}} & 
\multicolumn{1}{c}{\textbf{Avg.}} & \multicolumn{1}{c}{\textbf{Std.}} \\ 
\midrule
Train & 1,379,986 & 98,791 (7.16\%) & 65,235 (4.73\%) & 4,555 (0.33\%) & 1,211,405 (87.78\%) & 12.4 & 7.6 \\
Dev & 179,371 & 13,161 (7.34\%) & 7,544 (4.21\%) & 534 (0.3\%) & 158,132 (88.16\%) & 12.1 & 7.2 \\
Test (news) & 87,721 & 6,263 (7.14\%) & 4,102 (4.68\%) & 305 (0.35\%) & 77,051 (87.84\%) & 12.4 & 7.2 \\
Test (Ref.) & 6,821 & 996 (14.6\%) & 279 (4.09\%) & 170 (2.49\%) & 5,376 (78.82\%)  & 4.8 & 3.2 \\
Test (ASR) & 6,417 & 887 (13.82\%) & 253 (3.94\%) & 125 (1.95\%) & 5,152 (80.29\%) & 5.3 & 3.6 \\ 
\bottomrule
\end{tabular}
}

\end{table}

%% file: sections/experiments/mt.tex
\subsection{Machine Translations: Bangla to English}
\label{ssec:exp_mt}


\subsubsection{Dataset}
\label{sssec:dataset_mt}

We used publicly available datasets reported in \cite{hasan2019icbslp,hasan2019neural}. A brief detail of each dataset is discussed below.


\begin{enumerate}
    \item \textbf{Six Indian Parallel Corpora \cite{post2012constructing} (SIPC):} The SIPC consists of the corpora of six languages, and the data was collected from the top-100 most-viewed documents from the Wikipedia page of each language \cite{post2012constructing}. The corpora contain $\sim$20K, $914$, $1,000$ parallel sentences in the training, development and test set, respectively.
    
    \item \textbf{Open Subtitles \cite{lison2016opensubtitles2016}:} The Open Subtitles corpus was developed from the translations of movie subtitles. We used the recent version (v2018) of the said corpus, which consists of $\sim$413K parallel sentences.
    
    \item \textbf{Indic Languages Multilingual Parallel Corpus (ILMPC):} The ILMPC was developed in the Workshop in Asian Translation (WAT) 2018 \cite{Nakazawa2018}, and consists of seven parallel languages. The monolingual text of said corpus was collected from OPUS, which has translated for the WAT workshop. It consists of movies and subtitles of TV series. The original version of the corpus has a total $\sim337$K, $500$ and $1,000$ parallel sentences in the training, development, and test set, respectively. For the current study, we preprocessed and eliminated code-mixed sentences containing English words in Bangla sentences.
    
    \item \textbf{SUPara Corpus \cite{al2018suparaBenchmark,al2018suparaLatest}:} The SUpara corpus has been developed by Shahjalal University of Science and Technology (SUST), Bangladesh, in which data was translated from several categories of newspaper articles such as literature, journalistic, external communication, administrative, etc. The corpus has $\sim$70.8K parallel sentences for training, 500 sentences for each development and test set.
    
    \item \textbf{AmaderCAT \cite{hasan2020collaborative}:} The AmaderCAT dataset has been developed using an open-source platform named AmaderCAT. The corpus has a total of $1,782$ parallel sentences.
    
    \item \textbf{Penn Treebank Bangla-English parallel corpus (PTB):} The Bangladesh team of the PAN Localization Project\footnote{http://www.panl10n.net} developed the PTB English-Bangla corpus. The dataset has $1,313$ parallel sentences, in which English sentences were collected from the Penn Treebank corpus.
    
    \item \textbf{Global Voices:\footnote{\url{http://casmacat.eu/corpus/global-voices.html}}} The Global Voices corpus consists of the translations of spoken languages. We used the latest version (2018Q4) of this corpus, in which the amount of parallel segment is $\sim$137K.

    \item \textbf{Tatoeba:\footnote{\url{https://tatoeba.org/}}} The Tatoeba corpus has been developed using the Tatoeba open-source platform. The corpus has $\sim$5.1k parallel sentences for the Bangla-English language pair.
    
    \item \textbf{Tanzil:\footnote{\url{http://tanzil.net/}}} The Tanzil corpus is derived from the Tanzil Project of Quran translations. This dataset consists of $\sim$187K parallel sentences.
    
\end{enumerate}

In Table \ref{table:mt_corpus_stats}, we present the statistics of the datasets that we used in our study. We used a total of 1,162,504 parallel sentences in our training set, which consists of $\sim$15.4M Bangla and $\sim$15.1M English tokens. The distribution of training, development, and test splits is reported in Table \ref{table:data-split-mt}.

\begin{table}[h]
\centering
\caption{Statistics of the \textbf{MT} datasets. Bangla (BN), English (EN).}
\label{table:mt_corpus_stats}
\scalebox{0.9}{
\begin{tabular}{@{}lrr@{}}
\toprule
\multicolumn{1}{c}{\textbf{Corpus Name}} & \multicolumn{1}{c}{\textbf{\# of Sentences}} & \multicolumn{1}{c}{\textbf{\# of Tokens}} \\ \midrule
SUPara & 70,861 & 813,184 (BN), 995,255 (EN) \\
ILMPC & 324,366 & 2,247,958 (BN), 2,675,011 (EN) \\ 
SIPC & 20,788 & 263,122 (BN), 323,200 (EN) \\ 
Global Voices & 137,620 & 2,567,115 (BN), 2,858,694 (EN) \\
Open Subtitles & 413,602 & 2,573,874 (BN), 3,011,878 (EN) \\
Tatoeba & 5,120 & 27,705 (BN), 30,069 (EN) \\
Tanzil & 187,052 & 6,880,944 (BN), 5,185,136 (EN) \\
PTB & 1,313 & 31,511 (BN), 32,220 (EN) \\
AmaderCAT & 1,782 & 13,698 (BN), 19,356 (EN) \\ 
\bottomrule
\end{tabular}
}
\end{table}



\begin{table}[h]
\centering
\caption{Data splits and distribution for the \textbf{MT} dataset. BN-Bangla, EN-English}
\label{table:data-split-mt}
\scalebox{0.9}{
\begin{tabular}{lrrl}
\toprule
\multicolumn{1}{c}{\textbf{Data set}} & \multicolumn{1}{c}{\textbf{\# of Sent}} & \multicolumn{1}{c}{\textbf{\# of Token}} \\ \midrule
\textbf{Train} & 1,162,504 & 15,419,111 (BN), 15,127,504 (EN) \\ 
\textbf{Dev} & 500 & 8,742 (BN), 10,815 (EN) \\ 
\textbf{Test} & 500 & 8,699 (BN), 10,817 (EN) \\ \bottomrule
\end{tabular}
}
\end{table}

\subsubsection{Training}
\label{sssec:training_mt}
As a part of the training, we first preprocessed the data, which includes removing all the parallel sentences containing English words in Bangla sentences, followed by tokenization, and binarization. 
In terms of tokenization, we used the BNLP Toolkit\footnote{\url{https://github.com/sagorbrur/bnlp.git}} to tokenize Bangla sentences and the NLTK tokenizer for English sentences. We also used the subword-nmt toolkit \cite{sennrich2015neural} to segment the text into subword units and applied byte pair encoding to increase the consistency of data segmentation for handling rare words. Finally, we applied the fairseq~\cite{ott2019fairseq} prepossessing script to binarize the data for training.

We used the WMT transformer architecture of the fairseq toolkit~\cite{ott2019fairseq}\footnote{\url{https://github.com/pytorch/fairseq}} to train the dataset. The training hyper-parameters of the transformer include optimizer Adam, weight decay 0.0001, learning rate 5e-4,  learning rate scheduler inverse\_sqrt, dropout 0.3, and warmup updates 8000. For validation, we used a beam search with beam size 5. For the evaluation, we used the SUPara test set (version 2018), which consists of 500 parallel sentences.

%% file: sections/experiments/affective_behavior.tex
\subsection{Sentiment Classification}
\label{ssec:sentiment}

\subsubsection{Dataset}
\label{sssec:dataset_sentiment}
Compared to other tasks and datasets, the interest in sentiment analysis research has been significantly higher. Over time, several resources have been developed. For the current study, we used publicly available datasets reported in \cite{AridSentiment2020}, individual and consolidated versions. In the following, we provide a brief description for each dataset.

\begin{enumerate}
\item \textbf{Sentiment Analysis in Indian Languages (SAIL) Dataset \cite{patra2015shared}:} The SAIL dataset has been developed in the Shared task on Sentiment Analysis in Indian Languages (SAIL) 2015, which consists of posts from Twitter. The training, development, and test set of this dataset has 1000, 500, and 500 tweet posts. In our study, we only use the train set and split the train set into train, development, and test set. 
 
\item \textbf{ABSA Dataset \cite{rahman2018datasets}:} The ABSA dataset was developed to perform aspect-based sentiment analysis task in Bangla. The dataset contains two categories of data which are cricket and restaurant. In the cricket category, authors collected data from Facebook, BBC Bangla, and Prothom Alo and manually annotated them, and in the restaurant category, authors directly translated the English benchmark’s Restaurant dataset \cite{pontiki-etal-2014-semeval}.
 
\item \textbf{YouTube Comments Dataset \cite{tripto2018detecting}:} The YouTube comments dataset was developed by extracting comments from various YouTube videos. The dataset contains three-class and five-class sentiment annotation. In our study, we only took the data of three class labels and converted the five class into three class labels. As a result, we have a total of 2,796 comments which we split into training, development, and test set in order to run individual experiments.

\item \textbf{BengFastText Dataset \cite{rezaul2020classification}:} The BengFastText dataset was collected from several newspapers, TV news, books, blogs, and social media. The original dataset reports 320,000 instances; however, a fraction of it is publicly available.\footnote{\url{https://github.com/rezacsedu/Classification_Benchmarks\_Benglai\_NLP/}} The public version include 8,420 posts including a test set. First, we combined the train and test set, and then we split the data into training, development, and test set.

\item \textbf{Social Media Posts (CogniSenti Dataset) \cite{AridSentiment2020}:} The CogniSenti dataset consist of 942 posts from Facebook and 5,628 tweets from Twitter. In order to train our model, we split the data into training, development, and test set comprising 4,599, 985, and 986 data, respectively.
\end{enumerate}


For the classification experiments, we used the same data splits reported in \cite{AridSentiment2020}, in which the training, development, and test sets consist of 70\%, 15\%, and 15\% proportion, respectively. In Table \ref{tab:dataset-sentiment}, we report the distribution of the data splits. We also consolidated them to see if data consolidation helps, where all train sets are combined into a single train set. The same procedures are applied to combine development and test sets.  

\begin{table}[]
\centering
\caption{Data splits and distributions of \textbf{Sentiment Classification} datasets}
\label{tab:dataset-sentiment}
\setlength{\tabcolsep}{4pt}
\scalebox{0.8}{
\begin{tabular}{@{}lrrrr@{}}
\toprule
\multicolumn{1}{c}{\textbf{Class label}} & \multicolumn{1}{c}{\textbf{Train}} & \multicolumn{1}{c}{\textbf{Dev}} & \multicolumn{1}{c}{\textbf{Test}} & \multicolumn{1}{c}{\textbf{Total}} \\ \midrule
\multicolumn{5}{c}{\textbf{ABSA: Cricket Dataset}} \\ \midrule
Positive & 376 & 71 & 73 & 520 \\
Neutral & 194 & 27 & 34 & 255 \\
Negative & 1,515 & 274 & 273 & 2,062 \\\cline{2-5}
\textbf{Total} & 2,085 & 372 & 380 & 2,837 \\ \midrule
\multicolumn{5}{c}{\textbf{ABSA: Restaurant Dataset}} \\ \midrule
Positive & 872 & 143 & 116 & 1,131 \\
Neutral & 167 & 35 & 46 & 248 \\
Negative & 326 & 46 & 57 & 429 \\\cline{2-5}
\textbf{Total} & 1,365 & 224 & 219 & 1,808 \\ \midrule
\multicolumn{5}{c}{\textbf{BengFastText Dataset}} \\ \midrule
Positive & 2,403 & 595 & 788 & 3,786 \\
Negative & 3,107 & 783 & 744 & 4,634 \\\cline{2-5}
\textbf{Total} & 5,510 & 1,378 & 1,532 & 8,420 \\ \midrule
\multicolumn{5}{c}{\textbf{SAIL}} \\ \midrule
Positive & 193 & 27 & 57 & 277 \\
Neutral & 257 & 36 & 75 & 368 \\
Negative & 247 & 35 & 72 & 354 \\\cline{2-5}
\textbf{Total} & 697 & 98 & 204 & 999 \\ \midrule
\multicolumn{5}{c}{\textbf{Youtube Comments Dataset}} \\ \midrule
Positive & 553 & 103 & 96 & 752 \\
Neutral & 539 & 106 & 116 & 761 \\
Negative & 865 & 210 & 208 & 1,283 \\\cline{2-5}
\textbf{Total} & 1,957 & 419 & 420 & 2,796 \\ \midrule
\multicolumn{5}{c}{\textbf{Social Media Posts (CogniSenti Dataset)}} \\ \midrule
Positive & 1,047 & 205 & 236 & 1,488 \\
Neutral & 2,633 & 553 & 563 & 3,749 \\
Negative & 919 & 227 & 187 & 1,333 \\ \cline{2-5}
\textbf{Total} & 4,599 & 985 & 986 & 6,570 \\ 
\midrule
\multicolumn{5}{c}{\textbf{Combined dataset}} \\ \midrule
Positive & 5,444 & 1,144 & 1,366 & 7,954 \\
Neutral & 3,790 & 757 & 834 & 5,381 \\
Negative & 6,979 & 1,575 & 1,541 & 10,095 \\ \cline{2-5}
\textbf{Total} & 16,213 & 3,476 & 3,741 & 23,430 \\ 
\bottomrule
\end{tabular}
}
\end{table}


\subsubsection{Training}
\label{sssec:traning_sentiment}
Social media content is always noisy, consisting of many symbols, emoticons, URLs, usernames, and invisible characters. Previous studies show that filtering and cleaning the data before training a classifier helps significantly. Hence, we preprocess the data before classification experiments. The preprocessing steps included removing stop words, invisible characters, punctuations, URLs, and hashtag signs. For the experiments, we used different pre-trained transformer models discussed in Section \ref{sec:methods}. We followed a fine-tuning procedure using a task-specific layer on top of the transformers network.

\subsection{Emotion Classification}
\label{ssec:emotion}

\subsubsection{Dataset}
\label{sssec:dataset_emotion}

We used the emotion detection dataset reported in \cite{tripto2018detecting,alam2020bangla}, which has been collected from YouTube video comments. The videos are manually selected from different domains, including music, sports, drama, news, etc. The dataset contains 2890 youtube comments in Bangla, English, and romanized Bangla. The annotation of the dataset consists of five emotion labels such as {\em(i)} anger/disgust, {\em(ii)} joy, {\em(iii)} sadness, {\em(iv)} fear/surprise, and {\em(v)} none. For the experiments, we use the same data splits reported in \cite{alam2020bangla}
Specifically, we set aside 10\% of the dataset for testing. The rest was further divided into 80\% for training and 20\% for development. In Table \ref{tab:dataset-emotion}, we report class label distribution of the emotion dataset.


\begin{table}[]
\centering
\caption{Data splits and distributions of \textbf{Emotion Classification} dataset.}
\label{tab:dataset-emotion}
\setlength{\tabcolsep}{4pt}
\scalebox{0.8}{
\begin{tabular}{@{}lrrrr@{}}
\toprule
\multicolumn{1}{c}{\textbf{Class label}} & \multicolumn{1}{c}{\textbf{Train}} & \multicolumn{1}{c}{\textbf{Dev}} & \multicolumn{1}{c}{\textbf{Test}} & \multicolumn{1}{c}{\textbf{Total}} \\ \midrule
Anger{/}Disgust& 	623	& 164	& 93	& 880 \\
Joy	& 545	& 136	& 94	& 775 \\
Sadness	& 70	& 26	& 13	& 109 \\
Fear{/}Surprise	& 94	& 27	& 19	& 140 \\
None	& 535	& 148	& 67	& 750 \\
\cline{2-5}
\textbf{Total} & 1,867	& 501	& 286	& 2,654 \\  
\bottomrule
\end{tabular}
}
\end{table}

\subsubsection{Training}
\label{sssec:traning_emotion}
For the experiments, we trained different pre-trained transformer models discussed in Section \ref{sec:methods}. All models were trained for 10 epochs with a learning rate of 1e-5 and a sequence length of 30. We use 32 samples in each mini-batch, except when this does not fit in memory (e.g., XLM-RoBERTa). In such cases, we use a maximum batch size that fits within GPU memory. All model parameters were fine-tuned during training, i.e., no layer was kept frozen. The model with the best development set performance was evaluated on the test dataset.

%% file: sections/experiments/news_categorization.tex
\subsection{News Categorization}
\label{ssec:exp_news_cat}

\subsubsection{Dataset}
This dataset was prepared for the news categorization task in \cite{kunchukuttan2020ai4bharat,alam2020bangla}. It contains six different class labels and is available with training, development, and test splits with 11284, 1411, and 1411 news articles, respectively. The class distribution is presented in Table \ref{tab:dataset-news}, in which the dataset has low distribution for \textit{International} news category. 


\begin{table}[]
\centering
\caption{Data splits and distributions of \textbf{News Categorization} dataset.}
\label{tab:dataset-news}
\setlength{\tabcolsep}{4pt}
\scalebox{0.8}{
\begin{tabular}{@{}lrrrr@{}}
\toprule
\multicolumn{1}{c}{\textbf{Class label}} & \multicolumn{1}{c}{\textbf{Train}} & \multicolumn{1}{c}{\textbf{Dev}} & \multicolumn{1}{c}{\textbf{Test}} & \multicolumn{1}{c}{\textbf{Total}} \\ \midrule
Kolkata	& 4,596	& 596	& 569	& 5,761 \\
State	& 2,189	& 246	& 278	& 2,713 \\
National	& 1,408	& 179	& 175	& 1,762 \\
Sports	& 1,257	& 151	& 191	& 1,599 \\
Entertainment	& 1,157	& 166	& 130	& 1,453 \\
International	& 515	& 71	& 66	& 652 \\
\cline{2-5}
\textbf{Total} & 11,122	& 1,409	& 1,409	& 13,940 \\  
\bottomrule
\end{tabular}
}
\end{table}

\subsubsection{Training}
\label{sssec:traning_news_classification}

We trained the models using cross-entropy loss criterion, and the Adam optimization algorithm \cite{DBLP:journals/corr/KingmaB14}. All models were trained for 10 epochs with a learning rate of 1e-5. We use 32 samples in each mini-batch, except when this does not fit in memory (e.g., XLM-RoBERTa large model). 
We used a fixed sequence length during training and added padding or truncated length when necessary. The sequence length consists of 300 tokens. All model parameters are fine-tuned during training, i.e., no layer is kept frozen. The model with the best validation set performance was evaluated on the test dataset.


\subsection{Authorship Attribution}
\label{ssec:exp_authorship}

\subsubsection{Dataset}
\label{sssec:dataset_authorship}
We used the dataset reported in \cite{khatun2019authorship,alam2020bangla}, which contains writings of 14 different authors from an online Bangla e-library (e.g., novels, story, series, etc.). Each document in the dataset has a fixed length of 750 words. The dataset was balanced so that each author has the same number of samples. The data splits consist of 14,047, 3,511, and 750 writings for train, development, and test splits, respectively. The class label distribution is reported in Table \ref{tab:dataset-authorship}. 


\begin{table}[]
\centering
\caption{Data splits and distributions of \textbf{Authorship Attribution} dataset.}
\label{tab:dataset-authorship}
\setlength{\tabcolsep}{4pt}
\scalebox{0.8}{
\begin{tabular}{@{}lrrrr@{}}
\toprule
\multicolumn{1}{c}{\textbf{Class label}} & \multicolumn{1}{c}{\textbf{Train}} & \multicolumn{1}{c}{\textbf{Dev}} & \multicolumn{1}{c}{\textbf{Test}} & \multicolumn{1}{c}{\textbf{Total}} \\ \midrule
Humayun Ahmed	& 2,898	& 714	& 906	& 4,518 \\
Shunil Gongopaddhay	& 1,230	& 340	& 393	& 1,963 \\
Shomresh	& 911	& 215	& 282	& 1,408 \\
Shorotchandra	& 833	& 218	& 261	& 1,312 \\
Robindronath	& 808	& 199	& 252	& 1,259 \\
MZI	& 715	& 165	& 220	& 1,100 \\
Shirshendu	& 660	& 178	& 210	& 1,048 \\
Toslima Nasrin	& 605	& 140	& 186	& 931 \\
Shordindu	& 567	& 144	& 177	& 888 \\
Shottojit Roy	& 553	& 127	& 169	& 849 \\
Tarashonkor	& 500	& 120	& 155	& 775 \\
Bongkim	& 350	& 100	& 112	& 562 \\
Nihar Ronjon Gupta	& 305	& 76	& 95	& 476 \\
Manik Bandhopaddhay	& 302	& 74	& 93	& 469 \\
\cline{2-5}
\textbf{Total} & 11,237	& 2,810	& 3,511	& 17,558 \\  
\bottomrule
\end{tabular}
}
\end{table}

\subsubsection{Training}
\label{sssec:traning_authorship}
We balanced the dataset prior to training, taking a minimum number of samples (469) per class similar to \cite{khatun2019authorship}. We limited the sequence length to 300 on this dataset even though each sample in the dataset has 750 words. This was done to meet GPU memory constraints.

%% file: sections/results.tex
\section{Results}
\label{sec:results}

In this section, we report and discuss the results for each task. We report previous state-of-the-art results as baselines for which they are available and compare that with ours. The results that improve over the baseline are highlighted in bold form, and the best system is highlighted in bold and \ul{underlined}. For some tasks, an exact comparison was possible as we could use the same data splits.

\subsection{Parts of Speech}
\label{ssec:results_pos}

\begin{table}[t]
\centering
\caption{Results on LDC test set for \textbf{POS tagging}.}
\label{tab:result_pos_tagging}
\setlength{\tabcolsep}{4pt}
\scalebox{0.9}{
\begin{tabular}{@{}llrrrr@{}}
\toprule
\multicolumn{1}{c}{\textbf{Model}} & \multicolumn{1}{c}{\textbf{Train}} & \multicolumn{1}{c}{\textbf{Acc}} & \multicolumn{1}{c}{\textbf{P}} & \multicolumn{1}{c}{\textbf{R}} & \multicolumn{1}{c}{\textbf{F1}} \\ \midrule
Baseline (Token+CRFs)~\cite{alam2016bidirectional} & LDC & 42.4 & 25.7 & 29.9 & 70.1 \\
Baseline (BiLSTM-CRFs)~\cite{alam2016bidirectional} & LDC & 86.3 & 86.3 & 86.3 & 86.0\\
Bangla Electra & LDC & 79.0 & 73.4 & 71.1 & 72.2 \\
 & LDC + IITKGP & 80.2 & 74.8 & 72.7 & 73.7 \\
 & LDC + IITKGP + CRBLP Corpus & 80.4 & 75.1 & 73.0 & 74.1 \\
Indic-BERT & LDC & 88.3 & 84.7 & 84.2 & 84.5 \\
 & LDC + IITKGP & 87.9 & 84.2 & 83.9 & 84.1 \\
 & LDC + IITKGP + CRBLP Corpus & 87.7 & 84.0 & 83.7 & 83.9 \\
Indic-DistilBERT & LDC & 89.8 & 86.7 & 86.5 & \textbf{86.6} \\
 & LDC + IITKGP & 89.6 & 86.3 & 86.0 & \textbf{86.2} \\
 & LDC + IITKGP + CRBLP Corpus & 89.6 & 86.3 & 86.1 & \textbf{86.2} \\
Indic-RoBERTa & LDC & 87.0 & 83.1 & 82.1 & 82.6 \\
 & LDC + IITKGP & 86.4 & 82.5 & 81.5 & 82.0 \\
 & LDC + IITKGP + CRBLP Corpus & 86.7 & 82.9 & 82.0 & 82.5 \\
Indic-XLM-RoBERTa & LDC & 87.7 & 83.8 & 83.4 & 83.6 \\
 & LDC + IITKGP & 87.7 & 83.8 & 83.7 & 83.7 \\
 & LDC + IITKGP + CRBLP Corpus & 87.4 & 83.5 & 83.2 & 83.4 \\
BERT-bn & LDC & 85.8 & 81.5 & 80.7 & 81.1 \\
 & LDC + IITKGP & 85.8 & 81.5 & 80.9 & 81.2 \\
 & LDC + IITKGP + CRBLP Corpus & 85.4 & 81.0 & 80.2 & 80.6 \\
BERT-m & LDC & 88.1 & 84.2 & 83.8 & 84.0 \\
 & LDC + IITKGP & 87.7 & 83.6 & 83.5 & 83.6 \\
 & LDC + IITKGP + CRBLP Corpus & 87.6 & 83.6 & 83.4 & 83.5 \\
DistilBERT-m & LDC & 83.9 & 78.8 & 78.0 & 78.4 \\
 & LDC + IITKGP & 83.7 & 78.5 & 77.9 & 78.2 \\
 & LDC + IITKGP + CRBLP Corpus & 83.8 & 78.4 & 78.0 & 78.2 \\
XLM-RoBERTa & LDC & 90.1 & 87.0 & 86.4 & \ul{\textbf{86.7}} \\
 & LDC + IITKGP & 89.6 & 86.2 & 86.1 & \textbf{86.2} \\
 & LDC + IITKGP + CRBLP Corpus & 89.5 & 86.1 & 86.0 & \textbf{86.1} \\ \bottomrule
\end{tabular}
}
\end{table}

For POS tagging experiments, we used nine pre-trained transformer models and trained the models using different combinations of the LDC, IITKGP, and CRBLP corpora. More specifically, we combined the IITKGP and CRBLP corpora with the LDC training set to enlarge the training data. In Table \ref{tab:result_pos_tagging}, we present the performance of the models. From the results, we observe that having additional data did not help improve the models' performance. In comparing our results with previous state-of-the-art work, we observe that there is only 0.6\% absolute improvement, which is considerably small. We investigated a previous study \cite{alam2016bidirectional} and realized that the model had been developed using a lexicon combining training, development, and test datasets, which may result in an overestimated performance. Hence, the current results are not exactly comparable with previous results. Comparing the models, Indic-DistilBERT and XLM-RoBERTa perform similarly and show higher performance compared to other models.
In comparing monolingual and multilingual models, XLM-RoBERTa shows higher performance, which might be because it is a large version of the model, whereas monolingual models are the base version, consisting of fewer parameters than the larger version. Among the monolingual models, Electra is the worst performing model.



\subsection{Lemmatization}
\label{ssec:results_lemma}

In Table \ref{tab:result_lemmatization}, we present the results of the lemmatization task using nine transformer models. The baseline results \cite{chakrabarty-etal-2017-context} reported using accuracy, hence, we can compare the results using rhe accuracy metric. Our best model (i.e., XLM-RoBERTa) achieves 1\% absolute improvement in accuracy compared to the baseline. For this task, \textbf{BERT-bn} is the second-best model after XLM-RoBERTa in terms of accuracy and F1 measures. Indic-DistilBERT did not perform well for this task.    

\begin{table}[]
\centering
\caption{Results on the test set for \textbf{Lemmatization}.}
\label{tab:result_lemmatization}
\setlength{\tabcolsep}{4pt}
\scalebox{0.9}{
\begin{tabular}{@{}lrrrr@{}}
\toprule
\multicolumn{1}{c}{\textbf{Model}} & \multicolumn{1}{c}{\textbf{Acc}} & \multicolumn{1}{c}{\textbf{P}} & \multicolumn{1}{c}{\textbf{R}} & \multicolumn{1}{c}{\textbf{F1}} \\ \midrule
Baseline~\cite{chakrabarty-etal-2017-context} & 74.1 & - & - & - \\
Bangla Electra &  7.5 & 14.1 & 6.3 & 8.7  \\
Indic-BERT & 60.9 & 60.7 & 60.0 & 60.4 \\
Indic-DistilBERT & 56.1 & 56.0 & 55.3 & 55.6 \\
Indic-RoBERTa & 59.2 & 59.1 & 58.8 & 58.9 \\
Indic-XLM-RoBERTa & 61.3 & 61.0 & 60.0 & 60.5 \\
BERT-bn & 66.1 & 66.0 & 65.8 & 65.9 \\
BERT-m & 62.1 & 62.2 & 62.0 & 62.1 \\
DistilBERT-m & 60.7 & 60.7 & 60.3 & 60.5  \\
XLM-RoBERTa & \ul{\textbf{75.1}} & 75.1 & 74.9 & \ul{\textbf{75.0}} \\
\bottomrule
\end{tabular}
}
\end{table}


\subsection{Named Entity Recognition}
\label{ssec:results_ner}

In Table \ref{tab:result_ner}, we report the results for NER. For this task, we experiment with different types of input, i.e., token only and with additional inputs. With token-only experiments, all models show a consistent improvement compared to the baseline \cite{chowdhury2018towards} except the Bangla Electra model. For NER, the literature shows that POS and GZ helps in improving the performance \cite{alam2012ner,alam2013combination}; and so we used them in our experiments. Adding them with the transformer models significantly improved performance. Note that the results in ~\cite{chowdhury2018towards} are reported in a partial and exact match of entities where we used exact match for the current study and compare the same with previous results. Similar to other tasks, XLM-RoBERTa performs better than other monolingual and multilingual models. For token-only experiments, multilingual BERT is the second best model, whereas, for \textit{Token + POS + GZ}, Indic-BERT is the second best model.  

\begin{table}[]
\centering
\caption{Results on the test set for \textbf{NER}. GZ: Gazetteers.}
\label{tab:result_ner}
\setlength{\tabcolsep}{4pt}
\scalebox{0.9}{
\begin{tabular}{@{}lrrrr@{}}
\toprule
\multicolumn{1}{c}{\textbf{Model}} & \multicolumn{1}{c}{\textbf{Acc}} & \multicolumn{1}{c}{\textbf{P}} & \multicolumn{1}{c}{\textbf{R}} & \multicolumn{1}{c}{\textbf{F1}} \\\midrule
\multicolumn{5}{c}{\textbf{Token}} \\\midrule
Baseline (Token)~\cite{chowdhury2018towards} & - & 48.0 & 34.0 & 40.0 \\
Bangla Electra & 68.6 & 27.3 & 16.7 & 20.7 \\
Indic-BERT & 80.6 & 46.9 & 58.6 & \textbf{52.1} \\
Indic-DistilBERT & 81.8 & 47.0 & 59.1 & \textbf{52.4} \\
Indic-RoBERTa & 78.3 & 39.1 & 47.8 & \textbf{43.0} \\
Indic-XLM-RoBERTa & 80.6 & 46.2 & 60.0 & \textbf{52.2} \\
BERT-bn & 82.5 & 47.3 & 56.4 & \textbf{51.4} \\
BERT-m & 81.9 & 50.2 & 59.0 & \textbf{54.3} \\
DistilBERT-m & 76.4 & 38.1 & 48.1 & \textbf{42.5}  \\
XLM-RoBERTa & 83.4 & 55.2 & 64.2 & \textbf{59.4} \\ \midrule
\multicolumn{5}{c}{\textbf{Token + POS   + GZ}} \\\midrule
Baseline (Token + POS)~\cite{chowdhury2018towards} & - & 65.0 & 53.0 & 58.0 \\
Baseline (Token (W2V) + POS + GZ)~\cite{chowdhury2018towards} & - & 56.0 & 56.0 & 56.0 \\
Bangla Electra & 76.8 & 35.3 & 38.5 & 36.8 \\
Indic-BERT & 85.6 & 57.5 & 68.4 & \textbf{62.5} \\
Indic-DistilBERT & 84.7 & 51.8 & 59.7 & \textbf{55.4} \\
Indic-RoBERTa & 84.7 & 51.8 & 59.7 & \textbf{55.4} \\
Indic-XLM-RoBERTa & 84.7 & 55.8 & 66.5 & \textbf{60.7} \\
BERT-bn & 83.1 & 49.4 & 59.8 & \textbf{54.1} \\
BERT-m & 84.8 & 57.2 & 64.0 & \textbf{60.4} \\
DistilBERT-m & 81.5 & 46.4 & 54.8 & \textbf{50.2} \\
XLM-RoBERTa & 87.0 & 63.6 & 70.5 & \ul{\textbf{66.9}}  \\ \bottomrule
\end{tabular}
}
\end{table}

\subsection{Punctuation Restoration}
\label{ssec:results_punc}

In Table \ref{tab:res-punc-res}, we report the results of the punctuation restoration task, which comprises news, manual, and ASR transcriptions. For this task, we report the precision ($P$), recall ($R$), and F1, and we present results for each punctuation category. The overall score was calculated by ignoring the no punctuation entries (\textit{O} tokens). We observed that monolingual models did not perform well for this task. The results reported in \cite{alam-etal-2020-punctuation} show that the XLM-RoBERTa large model performs the best across different datasets such as news, manual, and ASR transcriptions. Among the monolingual models, Indic-DistilBERT performs better overall. For news text, manual, and ASR transcriptions, the F1 score is 80\%, 61.5\%, and 58.0\%, respectively.  

As expected, the performance on the news test set is better than the transcribed texts for all models. Due to errors introduced by the ASR model, performance in ASR transcriptions are lower than manual transcriptions. Among different labels, performance in \textit{Comma} is significantly worse in the transcribed texts.

\begin{table}[t]
\centering
\caption{Result on News, Ref. and ASR test datasets for the \textbf{Punctuation restoration} task. For overall best results we use bold and underlined form.
}
\label{tab:res-punc-res}
\setlength{\tabcolsep}{4pt}
\scalebox{0.75}{
\begin{tabular}{@{\extracolsep{1pt}}llccc|ccc|ccc|ccc@{}}
\toprule
\multirow{2}{*}{\textbf{Test}} & \multirow{2}{*}{\textbf{Model}}  & \multicolumn{3}{c|}{\textit{\textbf{Comma}}} & \multicolumn{3}{c|}{\textit{\textbf{Period}}} & \multicolumn{3}{c|}{\textit{\textbf{Question}}} & \multicolumn{3}{c}{\textit{\textbf{Overall}}}  \\
\cline{3-14} 
& & \textbf{P} & \textbf{R} & \textbf{F1}
& \textbf{P} & \textbf{R} & \textbf{F1}
& \textbf{P} & \textbf{R} & \textbf{F1}
& \textbf{P} & \textbf{R} & \textbf{F1} \\
 \midrule
\multirow{12}{*}{\textbf{News}} 
& BERT-m \cite{alam-etal-2020-punctuation} & 79.8	 & 68.2	 & 73.5	 & 80.4	 & 85.4	 & 82.8	 & 72.1	 & 77.0	 & 74.5	 & 79.9	 & 78.5	 & 79.2 \\
& DistilBERT-m \cite{alam-etal-2020-punctuation} & 72.1	 & 60.8	 & 66.0	 & 74.5	 & 71.6	 & 73.0	 & 56.9	 & 67.5	 & 61.8	 & 73.0	 & 67.3	 & 70.1 \\
& XLM-MLM-100-1280 \cite{alam-etal-2020-punctuation}& 76.9 & 71.2  & 73.9  & 82.0 &  83.4  & 82.9  & 70.2  & 76.4  & 73.2  & 80.0  & 78.5  & 79.3 \\
& {\cellcolor[HTML]{DAE8FC}}XLM-RoBERTa \cite{alam-etal-2020-punctuation}& {\cellcolor[HTML]{DAE8FC}}86.0 & {\cellcolor[HTML]{DAE8FC}}77.0  & {\cellcolor[HTML]{DAE8FC}}81.2  & {\cellcolor[HTML]{DAE8FC}}89.4  & {\cellcolor[HTML]{DAE8FC}}92.3  & {\cellcolor[HTML]{DAE8FC}}90.8  & {\cellcolor[HTML]{DAE8FC}}77.4  & {\cellcolor[HTML]{DAE8FC}}85.6  & {\cellcolor[HTML]{DAE8FC}}81.3  & {\cellcolor[HTML]{DAE8FC}}87.8  & {\cellcolor[HTML]{DAE8FC}}86.2  & {\cellcolor[HTML]{DAE8FC}}\ul{\textbf{87.0}} \\
\cline{2-14}
& Bangla Electra	& 66.9	& 30.4	& 41.8	& 64.2	& 64.6	& 64.4	& 60.0	& 1.0	& 1.9	& 64.8	& 49.7	& 56.3  \\
& Indic-BERT	& 70.4	& 63.6	& 66.8	& 76.3	& 75.8	& 76.1	& 66.7	& 53.8	& 59.5	& 73.9	& 70.5	& 72.2  \\
& Indic-DistilBERT	& 80.0	& 69.9	& 74.6	& 82.1	& 85.4	& 83.7	& 75.0	& 67.9	& 71.3	& 81.2	& 79.0	& 80.0  \\
& Indic-RoBERTa	& 73.5	& 59.8	& 66.0	& 76.7	& 74.5	& 75.6	& 60.8	& 61.6	& 61.2	& 75.1	& 68.5	& 71.7  \\
& Indic-XLM-RoBERTa	& 71.7	& 58.9	& 64.7	& 74.4	& 75.0	& 74.7	& 69.5	& 53.8	& 60.6	& 73.3	& 68.2	& 70.7  \\
& BERT-m	& 71.9	& 52.8	& 60.9	& 72.5	& 70.9	& 71.7	& 58.0	& 56.1	& 57.0	& 71.9	& 63.5	& 67.4  \\
\midrule
\multirow{12}{*}{\textbf{Ref.}}  
& BERT-m  \cite{alam-etal-2020-punctuation}& 35.6 & 34.4 & 35.0 & 67.4 & 64.7 & 66.0 & 39.8 & 28.8 & 33.4 & 58.5 & 54.6 & 56.5 \\
& DistilBERT-m \cite{alam-etal-2020-punctuation}& 32.6 & 31.5 & 32.1 & 64.0 & 50.2 & 56.3 & 32.5 & 14.7 & 20.2 & 54.3 & 42.4 & 47.6 \\
& XLM-MLM-100-1280 \cite{alam-etal-2020-punctuation}& 33.4 & 39.8 & 36.3 & 70.3 & 64.0 & 67.0 & 42.4 & 22.9 & 29.8 & 59.2 & 54.5 & 56.7 \\
& {\cellcolor[HTML]{DAE8FC}}XLM-RoBERTa \cite{alam-etal-2020-punctuation}& {\cellcolor[HTML]{DAE8FC}}39.3 & {\cellcolor[HTML]{DAE8FC}}36.9 & {\cellcolor[HTML]{DAE8FC}}38.1 & {\cellcolor[HTML]{DAE8FC}}76.9 & {\cellcolor[HTML]{DAE8FC}}81.4 & {\cellcolor[HTML]{DAE8FC}}79.1 & {\cellcolor[HTML]{DAE8FC}}54.3 & {\cellcolor[HTML]{DAE8FC}}58.8 & {\cellcolor[HTML]{DAE8FC}}56.5 & {\cellcolor[HTML]{DAE8FC}}67.6 & {\cellcolor[HTML]{DAE8FC}}70.2 & {\cellcolor[HTML]{DAE8FC}}\ul{\textbf{68.8}} \\
\cline{2-14}
& Bangla Electra	& 30.6	& 22.6	& 26.0	& 67.4	& 50.4	& 57.7	& 100.0	& 0.6	& 1.2	& 59.5	& 39.2	& 47.2    \\
& Indic-BERT	& 34.8	& 34.1	& 34.4	& 68.5	& 66.0	& 67.2	& 52.6	& 17.6	& 26.4	& 60.7	& 54.1	& 57.2    \\
& Indic-DistilBERT	& 39.5	& 32.3	& 35.5	& 72.1	& 71.9	& 72.0	& 54.2	& 18.8	& 27.9	& 65.5	& 58.0	& 61.5    \\
& Indic-RoBERTa	& 33.4	& 34.8	& 34.1	& 65.3	& 55.2	& 59.8	& 35.0	& 21.2	& 26.4	& 55.3	& 47.3	& 51.0    \\
& Indic-XLM-RoBERTa	& 36.6	& 32.3	& 34.3	& 67.7	& 66.5	& 67.1	& 47.4	& 15.9	& 23.8	& 60.8	& 53.9	& 57.2    \\
& BERT-bn	&34.5	&31.9	&33.1 & 68.4	&53.0	&59.7	&35.4	&20.6	&26.0	&57.8	&45.1	&50.7   \\
\midrule
\multirow{12}{*}{\textbf{ASR}}  
& BERT-m \cite{alam-etal-2020-punctuation}& 29.3 & 30.0 & 29.7 & 60.6 & 60.2 & 60.4 & 36.1 & 38.4 & 37.2 & 51.7 & 52.0 & 51.9 \\
& DistilBERT-m \cite{alam-etal-2020-punctuation}& 29.0 & 33.6 & 31.1 & 62.6 & 50.6 & 	56.0 & 31.3 & 20.8 & 25.0 & 51.2 & 44.3 & 47.5 \\
& XLM-MLM-100-1280 \cite{alam-etal-2020-punctuation}& 31.2 & 38.7 & 34.6 & 63.4 & 59.5 & 61.4 & 32.0 & 24.8 & 27.9 & 52.8 & 51.9 & 52.4 \\
& {\cellcolor[HTML]{DAE8FC}}XLM-RoBERTa \cite{alam-etal-2020-punctuation}& {\cellcolor[HTML]{DAE8FC}}38.3 & {\cellcolor[HTML]{DAE8FC}}35.6 & {\cellcolor[HTML]{DAE8FC}}36.9 & {\cellcolor[HTML]{DAE8FC}}69.2 & {\cellcolor[HTML]{DAE8FC}}77.2 & {\cellcolor[HTML]{DAE8FC}}73.0 & {\cellcolor[HTML]{DAE8FC}}38.5 & {\cellcolor[HTML]{DAE8FC}}52.0 & {\cellcolor[HTML]{DAE8FC}}44.2 & {\cellcolor[HTML]{DAE8FC}}60.3 & {\cellcolor[HTML]{DAE8FC}}66.4 & {\cellcolor[HTML]{DAE8FC}}\ul{\textbf{63.2}} \\
\cline{2-14}
& Bangla Electra	& 31.6	& 24.1	& 27.4	& 61.4	& 48.1	& 53.9	& 33.3	& 0.8	& 1.6	& 54.8	& 38.7	& 45.3  \\
& Indic-BERT	& 30.6	& 32.4	& 31.5	& 64.4	& 63.9	& 64.1	& 45.2	& 22.4	& 29.9	& 55.9	& 53.5	& 54.7  \\
& Indic-DistilBERT	& 38.1	& 32.8	& 35.2	& 64.9	& 69.1	& 67.0	& 46.8	& 17.6	& 25.6	& 59.4	& 56.8	& 58.0  \\
& Indic-RoBERTa	& 28.3	& 31.6	& 29.9	& 62.1	& 55.7	& 58.7	& 33.7	& 24.8	& 28.6	& 51.7	& 47.8	& 49.7  \\
& Indic-XLM-RoBERTa	& 28.8	& 31.2	& 30.0	& 62.0	& 63.8	& 62.9	& 44.7	& 16.8	& 24.4	& 54.0	& 52.6	& 53.3  \\
& BERT-bn	& 25.9	& 24.5	& 25.2	& 61.3	& 51.5	& 56.0	& 34.2	& 20.8	& 25.9	& 51.4	& 43.1	& 46.9  \\
\bottomrule
\end{tabular}
}
\end{table}

\subsection{Machine Translation}
\label{ssec:results_mt}
In Table \ref{table:result_nmt_supara_testset}, we present the performance of the transformer model, including baseline and state-of-art results on the SUPara test set. It is evident from the state-of-art results that our model provides the second best results on the SUPara test set.

\begin{table}[h]
\caption{Results on the SUPara test set for \textbf{MT}.}
\label{table:result_nmt_supara_testset}
\centering
\setlength{\tabcolsep}{4pt}
\scalebox{0.9}{
\begin{tabular}{@{}lr@{}}
\toprule
\multicolumn{1}{c}{\textbf{Experiments}} & \multicolumn{1}{c}{\textbf{BLEU}} \\ \midrule
shu-torjoma (baseline) \cite{mumin2019} & 17.4 \\ \midrule
BiLSTM \cite{hasan2019neural} & \textbf{19.9} \\
NMT \cite{mumin2019neural} & \textbf{22.7} \\
Transformer \cite{hasan2020not} & \textbf{32.1} \\ \midrule
Ours & \textbf{\textit{24.3}} \\ 
\bottomrule
\end{tabular}
}
\end{table}

The transformer-based models provide better results compared to other statistical and NMT models. However, it has problems with translating rare words, especially nouns. Increasing the amount of training data did not yield the performance expected, due to morphological complexity \cite{hasan2020collaborative}, and highly inflected words \cite{hasan2020not}. The translation quality of the test set is not up to the mark \cite{hasan2019neural,hasan2020not} and the training set contains both American and British English, which creates an impact on the overall performance. There is also translation from Indian Bangla to English, and some translations of lexemes differ from Bangladeshi Bangla (e.g., `jamuna', `yamuna').

\begin{figure}[t]
\centering
\includegraphics[width=0.60\columnwidth]{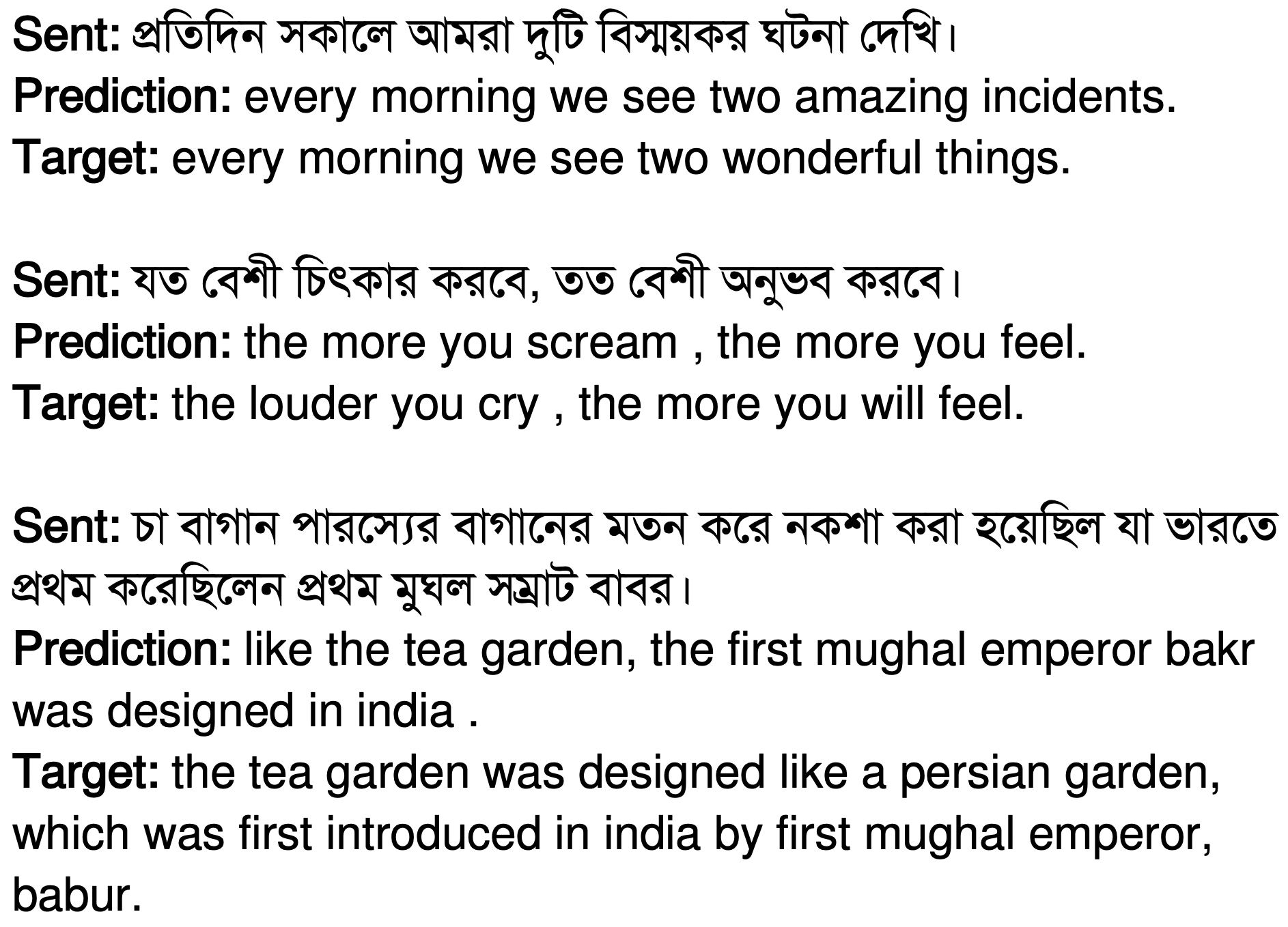}
\caption{Example sentences with wrong translations by transformer model.}
\label{fig:mt_example}  
\vspace{-0.4cm}
\end{figure}

In the first two sentences of Figure \ref{fig:mt_example}, we present examples of the incorrect translation of the target sentence and the appropriate prediction of the model. We observed that some of the words that have abbreviations (i.e., US Dollar and USD) have an effect on the translation score. In the third sentence of Figure \ref{fig:mt_example}, we present an example that our system could not predict correctly. These types of errors are mostly observed in sentences due to the presence of rare words.
Our study also reveals that the translation of a sentence might be entirely wrong in the presence of rare words or unknown words.


Another important issue that we observed is that the translations of spoken corpora differ from other corpora. For example, the Tanzil corpus also has translation issues, 
and is focused primarily on religious topics. The corpus has many long sentences, which made it difficult 
while training the transformer models. It also has more Bangla tokens than English compared to other corpora;
noisy words and covers different domains that create difficulties for the model to predict proper words for the target sequence. Another example is the GlobalVoices dataset, which creates difficulties in overall performance because of its long segments, and NMT can not translate long sentences properly \cite{koehn2017six}. 

Our reported results on the SUPara test set are the second-best existing system. However, reaching results that are on par with resource-rich languages is still a big challenge because of morphological complexity, namely inflectional morphemes and inflected words, which is further compounded by a lack of resources. More good quality translation is required to improve on these results for future work.

\subsection{Sentiment Classification}
\label{ssec:results_sentiment}

In Table~\ref{tab:results_sentiment_classification}, we report the classification results for each sentiment dataset. It summarizes the results obtained by different transformer models for individual and combined datasets. The study by Hasan et al. \cite{AridSentiment2020} report results on individual datasets, using classical and three transformers models such as the multilingual version of BERT-m, DistilBERT-m, and XLM-RoBERTa. They only report the F1 measure for each model. Here, we experiment using nine monolingual and multilingual transformer models for both individual and consolidated datasets. For all models, we report the accuracy, precision, recall, and F-measure. We had consistent improvement using the combined dataset across all models. It shows that data consolidation improves performance for the sentiment classification task. Among transformer-based models, BERT performs better on three datasets, and XLM-RoBERTa performs well on two datasets. For the combined dataset, XLM-RoBERTa is the best performing architecture. Indic-DistilBERT, Indic-BERT, and BERT-bn perform comparably on the combined dataset. 

\begin{table*}[ht]
\centering
\caption{Results on the test set for \textbf{Sentiment} classification. Best results on the combined dataset are highlighted with bold form.}
\label{tab:results_sentiment_classification}
\setlength{\tabcolsep}{4pt}
\scalebox{0.82}{
\begin{tabular}{lrrrrrrrrrrrr}
\toprule
\multicolumn{1}{c}{\textbf{}} & \multicolumn{4}{c}{\textbf{Bangla-Electra}} & \multicolumn{4}{c}{\textbf{Indic-BERT}} & \multicolumn{4}{c}{\textbf{Indic-DistilBERT}} \\\midrule
\multicolumn{1}{c}{\textbf{}} & \multicolumn{1}{c}{\textbf{Acc}} & \multicolumn{1}{c}{\textbf{P}} & \multicolumn{1}{c}{\textbf{R}} & \multicolumn{1}{c}{\textbf{F1}} & \multicolumn{1}{c}{\textbf{Acc}} & \multicolumn{1}{c}{\textbf{P}} & \multicolumn{1}{c}{\textbf{R}} & \multicolumn{1}{c}{\textbf{F1}} & \multicolumn{1}{c}{\textbf{Acc}} & \multicolumn{1}{c}{\textbf{P}} & \multicolumn{1}{c}{\textbf{R}} & \multicolumn{1}{c}{\textbf{F1}} \\ \midrule
ABSA cricket & 71.3 & 63.5 & 71.3 & 67.0 & 71.1 & 71.8 & 71.1 & 71.4 & 72.9 & 70.9 & 72.9 & 71.6 \\
ABSA restaurant & 50.2 & 35.7 & 50.2 & 40.9 & 60.3 & 58.7 & 60.3 & 58.7 & 65.8 & 65.3 & 65.8 & 64.7 \\
SAIL & 53.3 & 51.9 & 53.3 & 49.8 & 61.3 & 61.2 & 61.3 & 61.2 & 59.3 & 58.9 & 59.3 & 59.0 \\
BengFastText & 68.7 & 73.0 & 68.7 & 66.8 & 68.7 & 77.0 & 68.7 & 65.7 & 68.7 & 77.2 & 68.7 & 65.7 \\
Youtube comments & 67.4 & 66.2 & 67.4 & 66.6 & 75.7 & 75.0 & 75.7 & 75.2 & 74.8 & 75.1 & 74.8 & 74.9 \\
CogniSenti & 63.9 & 58.3 & 63.9 & 59.3 & 66.1 & 65.6 & 66.1 & 65.8 & 68.5 & 67.6 & 68.5 & 67.9 \\
\rowcolor[HTML]{DAE8FC}
Combined & 72.0 & 72.4 & 72.0 & 71.8 & 80.3 & 80.3 & 80.3 & 80.2 & 81.3 & 81.4 & 81.3 & 81.3 \\ \midrule
\multicolumn{1}{c}{\textbf{}} & \multicolumn{4}{c}{\textbf{Indic-RoBERTa}} & \multicolumn{4}{c}{\textbf{Indic-XLM-RoBERTa}} & \multicolumn{4}{c}{\textbf{BERT-bn}} \\ \midrule
ABSA cricket & 70.8 & 70.5 & 70.8 & 70.6 & 72.4 & 70.9 & 72.4 & 71.6 & 71.1 & 67.8 & 71.1 & 69.2 \\
ABSA restaurant & 68.0 & 67.2 & 68.0 & 66.9 & 63.5 & 63.4 & 63.5 & 62.1 & 60.7 & 59.3 & 60.7 & 58.5 \\
SAIL & 63.3 & 62.8 & 63.3 & 62.6 & 60.0 & 59.5 & 60.0 & 59.1 & 60.7 & 59.8 & 60.7 & 59.5 \\
BengFastText & 68.8 & 75.4 & 68.8 & 66.2 & 70.5 & 78.0 & 70.5 & 68.0 & 69.0 & 75.5 & 69.0 & 66.5 \\
Youtube comments & 75.0 & 74.8 & 75.0 & 74.8 & 74.3 & 74.5 & 74.3 & 74.3 & 73.3 & 72.6 & 73.3 & 72.9 \\
CogniSenti & 66.8 & 66.2 & 66.8 & 66.5 & 66.5 & 66.3 & 66.5 & 66.4 & 62.4 & 62.1 & 62.4 & 62.2 \\
\rowcolor[HTML]{DAE8FC}
Combined & 80.2 & 80.1 & 80.2 & 80.1 & 79.8 & 79.8 & 79.8 & 79.8 & 79.3 & 79.3 & 79.3 & 79.2 \\ \midrule
\multicolumn{1}{c}{\textbf{}} & \multicolumn{4}{c}{\textbf{BERT-m}} & \multicolumn{4}{c}{\textbf{DistilBERT-m}} & \multicolumn{4}{c}{\textbf{XLM-RoBERTa}} \\ \midrule
ABSA cricket & 67.7 & 69.2 & 67.7 & 68.2 & 74.5 & 66.5 & 74.5 & 69.9 & 67.7 & 68.2 & 67.7 & 68.2 \\
ABSA restaurant & 59.8 & 58.6 & 59.8 & 58.1 & 60.3 & 57.7 & 60.3 & 57.9 & 70.8 & 71.6 & 70.8 & 69.2 \\
SAIL & 56.7 & 56.6 & 56.7 & 56.6 & 57.3 & 57.0 & 57.3 & 57.0 & 56.7 & 56.6 & 56.7 & 56.6 \\
BengFastText & 69.6 & 75.6 & 69.6 & 67.4 & 69.1 & 74.6 & 69.1 & 66.9 & 69.6 & 75.3 & 69.6 & 67.4 \\
Youtube comments & 72.6 & 73.3 & 72.6 & 72.9 & 70.0 & 70.4 & 70.0 & 70.1 & 72.6 & 73.3 & 72.6 & 72.9 \\
CogniSenti & 68.9 & 68.5 & 68.9 & 68.6 & 59.2 & 58.6 & 59.2 & 58.9 & 68.9 & 68.5 & 68.9 & 68.6 \\
\rowcolor[HTML]{DAE8FC}
Combined & 80.0 & 79.9 & 80.0 & 79.9 & 74.3 & 74.8 & 74.3 & 74.2 & 82.0 & 82.0 & 82.0 & \ul{\textbf{82.0}} \\
\bottomrule
\end{tabular}
}
\end{table*}

\subsection{Emotion Classification}
\label{ssec:results_emotion}
In Table \ref{tab:results_emotion_classification}, we report the results for emotion classification and compare it to the performances reported in \cite{tripto2018detecting}. We obtained better results using the using XLM-RoBERTa model compared to others. Across multilingual transformer models, DistilBERT-m is the second best model. Among the monolingual models, the Indic-RoBERTa model performs better than other models for emotion classification.

\begin{table}[]
\centering
\caption{Results on the test set for \textbf{Emotion} classification.}
\label{tab:results_emotion_classification}
\setlength{\tabcolsep}{4pt}
\scalebox{0.82}{
\begin{tabular}{@{}lrrrr@{}}
\toprule
\multicolumn{1}{c}{\textbf{Model}} & \multicolumn{1}{c}{\textbf{Acc}} & \multicolumn{1}{c}{\textbf{P}} & \multicolumn{1}{c}{\textbf{R}} & \multicolumn{1}{c}{\textbf{F1}} \\ \midrule
Baseline (NB \cite{tripto2018detecting}) & 52.5 & - & - & 52.5 \\
Baseline (SVM \cite{tripto2018detecting}) & 49.3 & - & - & 49.8 \\
Baseline (CNN \cite{tripto2018detecting}) & 54.0 & - & - & 53.5 \\
Baseline (LSTM \cite{tripto2018detecting}) & 59.2 & - & - & 52.9 \\
Bangla Electra & 43.8 & 38.3 & 43.8 & 36.3 \\
Indic-BERT & 50.6 & 52.1 & 50.6 & 49.1 \\
Indic-DistilBERT & \textbf{59.4} & 58.4 & 59.4 & \textbf{57.3} \\
Indic-RoBERTa & \textbf{60.3} & 55.8 & 60.3 & \textbf{57.0} \\
Indic-XLM-RoBERTa & 53.1 & 48.8 & 53.1 & 49.6 \\
BERT-bn & 49.1 & 46.7 & 49.1 & 46.9 \\
BERT-m & \textbf{60.4} & 59.5 & 60.4 & \textbf{59.1} \\
DistilBERT-m & \textbf{69.8} & 68.4 & 69.8 & \textbf{66.6} \\
XLM-RoBERTa & \ul{\textbf{72.7}} & 70.3 & 72.7 & \ul{\textbf{70.6}} \\ \bottomrule
\end{tabular}%
}
\end{table}

\subsection{News Categorization}
\label{sec:results_news_categorization}
For the news categorization task, we report the results in Table \ref{tab:results_news_classification}. We used the same training, validation and test splits provided in \cite{kunchukuttan2020ai4bharat,alam2020bangla} for news categorization. 
The authors in \cite{kunchukuttan2020ai4bharat} used fastText embedding trained on different corpora to train the KNN classifier. In comparison, Alam et al. \cite{alam2020bangla} experiment with BERT-m and XLM-RoBERTa models. The performances of transformer models are higher than the results reported in \cite{kunchukuttan2020ai4bharat} in terms of accuracy. We obtained the best results using Indic-BERT, the second-best with Indic-XLM-RoBERTa and the next one is with XLM-RoBERTa. We can conclude from the results that transformer-based models are better suited for Bangla news categorization tasks than classical machine learning-based approaches that use manually engineered features or CNN/LSTM models trained on distributed word representation. 


\begin{table}[]
\centering
\caption{Results on the test set of \textbf{News} categorization.}
\label{tab:results_news_classification}
\setlength{\tabcolsep}{4pt}
\scalebox{0.82}{
\begin{tabular}{@{}lrrrr@{}}
\toprule
\multicolumn{1}{c}{\textbf{Model}} & \multicolumn{1}{c}{\textbf{Acc}} & \multicolumn{1}{c}{\textbf{P}} & \multicolumn{1}{c}{\textbf{R}} & \multicolumn{1}{c}{\textbf{F1}} \\ \midrule
Baseline (FT-W \cite{kunchukuttan2020ai4bharat}) & 62.8 & - & - &  \\
Baseline (FT-WC \cite{kunchukuttan2020ai4bharat}) & 64.8 & - & - &  \\
Baseline (INLP \cite{kunchukuttan2020ai4bharat}) & 72.5 & - & - &  \\
Bangla Electra & \textbf{80.4} & 78.5 & 80.4 & 79.2 \\
Indic-BERT & \ul{\textbf{94.1}} & 94.1 & 94.1 & \ul{\textbf{94.1}} \\
Indic-DistilBERT & \textbf{89.0} & 90.2 & 89.0 & 89.4 \\
Indic-RoBERTa & \textbf{81.6} & 81.6 & 81.6 & 81.4 \\
Indic-XLM-RoBERTa & \textbf{94.0} & 94.1 & 94.0 & 94.0 \\
BERT-bn & \textbf{91.0} & 91.2 & 91.0 & 91.0 \\
BERT-m & \textbf{91.3} & 91.5 & 91.3 & 91.3 \\
DistilBERT-m & \textbf{79.5} & 79.4 & 79.5 & 79.0 \\
XLM-RoBERTa & \textbf{93.4} & 93.4 & 93.7 & 93.4 \\ \bottomrule
\end{tabular}%
}
\end{table}



\begin{table}[tbh!]
\centering
\caption{Results on test set for \textbf{Authorship} classification.}
\label{tab:results_authorship_classification}
\setlength{\tabcolsep}{4pt}
\scalebox{0.8}{
\begin{tabular}{@{}lrrrr@{}}
\toprule
\multicolumn{1}{c}{\textbf{Model}} & \multicolumn{1}{c}{\textbf{Acc}} & \multicolumn{1}{c}{\textbf{P}} & \multicolumn{1}{c}{\textbf{R}} & \multicolumn{1}{c}{\textbf{F1}} \\ \midrule
Baseline (Char-CNN \cite{khatun2019authorship}) & 69.0 & - & - & - \\
Baseline (W2V (CBOW) \cite{khatun2019authorship}) & 71.8 & - & - & - \\
Baseline (fastText (CBOW) \cite{khatun2019authorship}) & 40.3 & - & - & - \\
Baseline (W2V (Skip) \cite{khatun2019authorship}) & 78.6 & - & - & - \\
Baseline (fastText (Skip) \cite{khatun2019authorship}) & 81.2 & - & - & - \\
Bangla Electra & 25.1 & 29.3 & 25.1 & 19.8 \\
Indic-BERT & \ul{\textbf{95.2}} & 95.3 & 95.2 & \ul{\textbf{95.2}} \\
Indic-DistilBERT & \textbf{87.4} & 87.6 & 87.4 & 87.3 \\
Indic-RoBERTa & 37.3 & 43.4 & 37.3 & 33.3 \\
Indic-XLM-RoBERTa & \textbf{94.6} & 94.8 & 94.6 & 94.6 \\
BERT-bn & \textbf{90.2} & 90.3 & 90.2 & 90.2 \\
BERT-m & \textbf{82.6} & 85.1 & 82.6 & 82.8 \\
DistilBERT-m & 58.8 & 58.6 & 58.6 & 57.6 \\
XLM-RoBERTa & \textbf{93.8} & 94.1 & 93.8 & 93.8 \\ \bottomrule
\end{tabular}%
}
\end{table}

\subsection{Authorship Attribution} 
\label{ssec:results_authorship}
In Table \ref{tab:results_authorship_classification}, we report the results with a comparison with previous work. The authors in \cite{khatun2019authorship} reported results for character and word level CNN models trained with fastText and word2vec embedding, and the best results were obtained with skip-gram variants of fastText embedding. All transformer models perform better than previously reported results except Bangla Electra, Indic-RoBERTa, and DistilBERT-m. We obtained the best results using the Indic-BERT model, accuracy of 95.2\%, which is a 14\% absolute improvement compared to the previous best model reported in \cite{khatun2019authorship}.

%% file: sections/challenges_future_work.tex
\section{Discussions, Challenges and Future Work}
\label{sec:challenge_future_work}

\subsection{Resources}

Some of the major challenges that are limiting further research in BNLP are: \textit{(i)} the availability and accessibility of resources; \textit{(ii)} the lack of reliability on annotated labels, with no proper guidelines or evaluation measures like inter-annotator agreement; and \textit{(iii)} the lack of comparable benchmarking performance -- due to the absence of well-defined train/dev/test splits.


To overcome these challenges, in Table \ref{tab:available_datasets}, we present the proprietary and publicly available -- 25 -- datasets for different tasks, listing their sizes in terms of the number of sentences and tokens; their content type (e.g., news articles, blogs, social media); license and URL. 
From the table, we can observe that there is a higher number of datasets for POS, MT, and sentiment tasks than others. In many cases, license information and train/dev/test splits are not available. It is also interesting to see that some datasets have been released as creative commons BY-SA type license, which allows others to use them for any purposes freely.

\subsection{Modeling}

\subsubsection{Model Comparison}
From our extensive literature review, we observed that the research has been following a trend from rule-based systems to deep learning-based approaches for different tasks. The recently developed transformer-based models have not been explored very extensively. Hence, we benchmark these tasks with transformer-based architecture in our study. We experiment with monolingual and multilingual transformer models, which also vary in their computational complexity (e.g., number of parameters). Across different tasks and pretrained models, we observed that the performances of different models vary. In Table \ref{tab:model_task_comparison}, we report the models with the top three ranks for eight different tasks.\footnote{We do not report MT tasks in the aforesaid Table as the MT model has been trained using one model.} Out of eight tasks in said Table, multilingual XLM-RoBERTa performs best in six tasks, whereas the Indic-BERT monolingual model performs best in two tasks. Indic-DistilBERT and Indic-XLM-RoBERTa are the second-best models. Each of them performed second best in three tasks. 

We hypothesize that the performances of the models are associated with their configuration in the architectural design. As shown in Table \ref{tab:model-stats}, XLM-RoBERTa is the largest in terms of vocab size, hidden units, number of layers, and attention heads; hence, this might be a reason that task-specific model trained using this architecture and pretrained model performed better for the majority cases. 

\begin{table}[]
\centering
\caption{Dataset for different NLP tasks. SIPC: Six Indian Parallel Corpora, YT: Youtube, News Cat.: News Categorization.}
\setlength{\tabcolsep}{4pt}
\scalebox{0.65}{
\begin{tabular}{@{}lllllll@{}}
\toprule
\multicolumn{1}{c}{\textbf{Task}} & \multicolumn{1}{c}{\textbf{Dataset}} &  \multicolumn{1}{c}{\textbf{Size}} & \multicolumn{1}{c}{\textbf{Content Type}} & \multicolumn{1}{c}{\textbf{License}} & \multicolumn{1}{c}{\textbf{Data Split}} & \multicolumn{1}{c}{\textbf{URL}} \\ \midrule
\multirow{3}{*}{POS} & \begin{tabular}[c]{@{}l@{}}Indian Language \\Part-of-Speech \\Tagset: Bengali \cite{baskaran2008common,baliLDC2010T16}\end{tabular} & \begin{tabular}[c]{@{}l@{}}7,398 sentences\\ 102,920 tokens\end{tabular} & \begin{tabular}[c]{@{}l@{}}Blogs, Wikipedia,\\ Multikulti and \\ a portion of the \\EMILLE/CIIL corpus\end{tabular} & LDC & \begin{tabular}[l]{@{}l@{}}train 70\%, \\dev 15\%, \\test 15\%\end{tabular} & \tablefootnote{\url{https://catalog.ldc.upenn.edu/LDC2010T16}} \\
& IITKGP Corpus \cite{iitgpgpostagging} & \begin{tabular}[c]{@{}l@{}}5,473 sentences, \\ 72,400 tokens\end{tabular} & News articles & NA & NA & NA \\
& CRBLP Corpus\cite{ummi2008developing} & \begin{tabular}[c]{@{}l@{}}1,176 sentences \\ 20,000 tokens\end{tabular} & News articles & NA & NA & NA \\ \midrule
Lemma & BenLem Dataset \cite{chakrabarty2017context} & 1,702 sentences & \begin{tabular}[c]{@{}l@{}}Rabindranath Tagore \\short stories \\and news articles\end{tabular} & NA & \begin{tabular}[l]{@{}l@{}}train 70\%, \\dev 15\%, \\test 15\%\end{tabular} & \tablefootnote{\url{https://www.isical.ac.in/utpal/docs/Bengali\_Dataset.rar}} \\ \midrule
NER & BCAB \cite{chowdhury2018towards} & 2,137 sentences & News articles & Proprietary & \begin{tabular}[l]{@{}l@{}}train 70\%, \\dev 10\%, \\test 20\%\end{tabular} & NA \\ \midrule
\begin{tabular}[c]{@{}l@{}}Punc.\end{tabular} & \begin{tabular}[c]{@{}l@{}}Punctuation \\Restoration \cite{alam-etal-2020-punctuation}\end{tabular} & \begin{tabular}[c]{@{}l@{}}4,700 news articles\\ 1.628M tokens\end{tabular} & \begin{tabular}[c]{@{}l@{}}News articles\end{tabular} & \multirow{2}{*}{CC BY 4.0, MIT} & \begin{tabular}[c]{@{}l@{}}train 1.38M,\\ dev 180K, \\test 88K tokens\end{tabular} & \tablefootnote{\url{https://github.com/xashru/punctuation-restoration}} \\
 &  & \begin{tabular}[c]{@{}l@{}}Manual trans. 6,821 tokens \\ ASR trans. 6,417 tokens\end{tabular} & Bangla short stories &  &  &  \\ \midrule
\multirow{9}{*}{MT} & SIPC \cite{post2012constructing} & 20,788 sentences &  & CC BY-SA 3.0 & NA & \tablefootnote{\url{https://github.com/joshua-decoder/indian-parallel-corpora}} \\
& OpenSubtitles \cite{lison2016opensubtitles2016} & 413,602 sentences & Movie subtitles & NA & NA & \tablefootnote{\url{https://www.opensubtitles.org/}} \\
& ILMPC \cite{Nakazawa2018} & 337K sentences & \begin{tabular}[c]{@{}l@{}}Movie subtitles, \\ Speech translation\end{tabular} & NA & NA & \tablefootnote{\url{http://lotus.kuee.kyoto-u.ac.jp/WAT/indic-multilingual/}} \\
& SUPara \cite{al2018suparaBenchmark,al2018suparaLatest} &  70,861 sentences &  & CC BY-SA 4.0 & \begin{tabular}[l]{@{}l@{}}train: 1,162,504,\\ dev: 500, test 500\\ sentences\end{tabular} & \tablefootnote{\url{https://ieee-dataport.org/open-access/supara-benchmark-benchmark-dataset-english-bangla-machine-translation}} \\
& AmaderCAT \cite{hasan2020collaborative} &  1,782 sentences & News articles & GPL-3.0 & NA & \tablefootnote{\url{https://github.com/AridHasan/Data-Collection-System-for-Machine-Translation}} \\
& PTB & 1,313 sentences & News articles &  LDC & NA &  \\
& GlobalVoices & 137,620 sentence & \begin{tabular}[c]{@{}l@{}}Speech corpora \\translation\end{tabular} & NA & NA & \tablefootnote{\url{https://opus.nlpl.eu/GlobalVoices-v2018q4.php}} \\
& Tatoeba & 5,120 sentences &  & CC BY 2.0 FR & NA & \tablefootnote{\url{https://opus.nlpl.eu/Tatoeba-v2021-03-10.php}} \\
& Tanzil & 187,052 sentences &  & Non-commercial & NA & \tablefootnote{\url{https://opus.nlpl.eu/Tanzil-v1.php}} \\ \midrule
\multirow{6}{*}{Sentiment} & ABSA Cricket \cite{rahman2018datasets} & 2,837 comments & \begin{tabular}[c]{@{}l@{}}FB posts, \\ News comments\end{tabular} & CC BY 4.0 & NA & \tablefootnote{\url{https://github.com/AtikRahman/Bangla\_ABSA\_Datasets}\label{absa_data}} \\
& ABSA Restaurant \cite{rahman2018datasets} &  1,808 translated sentences & Reviews & CC BY 4.0 & NA & {\footnotesize \ref{absa_data}} \\
& BengFastText \cite{rezaul2020classification} & 8,420 posts & \begin{tabular}[c]{@{}l@{}}News articles\\ FB posts\end{tabular} & NA & NA & \tablefootnote{\url{https://github.com/rezacsedu/Classification\_Benchmarks\_Benglai\_NLP}} \\
& SAIL \cite{patra2015shared} & 998 tweets & Twitter & Non-profit & \begin{tabular}[c]{@{}l@{}}train 70\%, \\dev 15\%, \\test 15\% \end{tabular}& \tablefootnote{\url{http://amitavadas.com/SAIL/}} \\
& YT Comments \cite{tripto2018detecting,AridSentiment2020} & 2,796 comments & Youtube & NA & \begin{tabular}[c]{@{}l@{}}train 70\%, \\dev 15\%, \\test 15\%\end{tabular} & \tablefootnote{\url{https://www.kaggle.com/nit003/bangla-youtube-sentiment-and-emotion-datasets}\label{youtube_sentiment}} \\
& CogniSenti & 6,570 posts & Twitter, FB Posts & Proprietary & \begin{tabular}[c]{@{}l@{}}train 70\%, \\dev 15\%, \\test 15\%\end{tabular} & NA \\ \midrule
Emotion & YT Emotion \cite{tripto2018detecting,alam2020bangla} & 2,890 comments & Youtube & NA & \begin{tabular}[c]{@{}l@{}}train 72\%, \\dev 18\%, \\test 10\%\end{tabular} & {\footnotesize \ref{youtube_sentiment}} \\ \midrule
News Cat. & News Corpus \cite{kunchukuttan2020ai4bharat,alam2020bangla} & 14,106 articles & News articles & CC BY-NC-SA 4.0  & \begin{tabular}[c]{@{}l@{}}train 80\%, \\dev 10\%, \\test 10\%\end{tabular} & \tablefootnote{\url{https://www.kaggle.com/csoham/classification-bengali-news-articles-indicnlp}} \\ \midrule
Authorship & Authorship Data \cite{khatun2019authorship,alam2020bangla} & 17,558 stories & E-book & CC BY 4.0 & \begin{tabular}[c]{@{}l@{}}train 77\%, \\ dev 19\% \\test 4\% \end{tabular}& \tablefootnote{\url{https://data.mendeley.com/datasets/6d9jrkgtvv/2}} \\ \bottomrule
\end{tabular}%
}
\label{tab:available_datasets}
\end{table}

\begin{table}[h]
\centering
\caption{Rank of the models across tasks. R$n$ represents the top three models in ranked order. Auth: Authorship, Emo: Emotion, Senti: Sentiment, Punc. : Punctuation Restoration. $**$ the model has been trained using a modified version of the original architecture.}
\label{tab:model_task_comparison}
\setlength{\tabcolsep}{4pt}
\scalebox{0.8}{
\begin{tabular}{@{}llccccccccc@{}}
\toprule
\multicolumn{1}{c}{\textbf{Lang}} & \multicolumn{1}{c}{\textbf{Model}} & \textbf{Model Type} & \textbf{POS} & \textbf{Lemma} & \textbf{NER} & \textbf{Punc.} & \textbf{Senti} & \textbf{Emo} & \textbf{News} & \textbf{Auth} \\ \midrule
\multirow{6}{*}{Mono} & Bangla Electra & base &  &  &  &  &  &  &  &  \\
 & Indic-BERT & base & R3 &  &  & R3 & R3 &  & R1 & R1 \\
 & Indic-DistilBERT & base & R2 &  &  & R2 & R2 &  &  &  \\
 & Indic-RoBERTa & $**$ &  &  &  &  &  &  &  &  \\
 & Indic-XLM-RoBERTa & $**$ &  &  & R2 &  &  &  & R2 & R2 \\
 & BERT-bn & base &  & R2 &  &  &  &  &  &  \\ \cmidrule(l){2-11} 
\multirow{3}{*}{Multi} & BERT-m & base &  & R3 & R3 &  &  & R3 &  &  \\
 & DistilBERT-m & base &  &  &  &  &  & R2 &  &  \\
 & XLM-RoBERTa & large & R1 & R1 & R1 & R1 & R1 & R1 & R3 & R3 \\ \bottomrule %
\end{tabular}%
}
\end{table}

\subsubsection{Inference Time Comparison:} While the said models are far better than classical algorithms (e.g., SVM, RF), it comes with significant computational complexity. Both the training and the inference time are significantly higher compared to classical algorithms. To give a better estimate, we compare the training and inference time between SVM and XLM-RoBERTa using the combined sentiment dataset (training set size 16,213 and test set size 3,741). 

The training and inference time for SVM was 2 hours 32 minutes 30 seconds, and 4 seconds, respectively. For XLM-RoBERTa, it took 3 hours 42 minutes 41 seconds for training and 1 minute 21 seconds for inference. Note that for SVM, we tune: C, G, kernel (linear and gamma) parameters. We fine-tune the XLM-RoBERTa for ten epochs. We run the experiments on a machine consisting of 22 CPUs, 50GB memory, and a GPU with 16GB memory. Such a finding clearly shows that a transformer-based model takes significantly high computation time even during inference. For example, in this case, the XLM-RoBERTa model took 20 times higher than SVM during inference. We obtained a weighted F1 score of 64.1\% using SVM and 82.0\% using XLM-RoBERTa, which is $\sim$18\% absolute improvement. This indicates that a trade-off needs to be taken into account for application deployment.

This is one of the significant challenges that the community is facing now. Our experiments show that performances vary across large vs. small models. Future research should also consider this aspect as this is important for real-time deployment.

\subsubsection{Error Analysis}
As the multilingual XLM-RoBERTa$_{large}$ model has performed better across different tasks, we analyzed its different characteristics including impact of tokenization to presence of different words and its learned representation. In the following, we discuss a number of key points in our findings from our analysis on several tasks. 

\begin{figure}[t]
\centering
\includegraphics[width=0.6\linewidth]{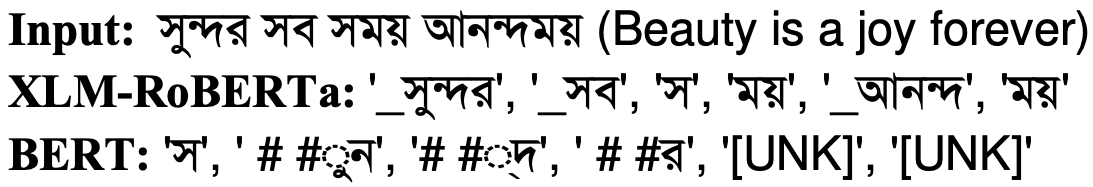}
\caption{Output of the different tokenizers.}
\label{fig:tranformer_tokenizers}  
\end{figure}

\textbf{Impact of Tokenization and Subword Embedding}: We realized that various factors are contributing to the performance gain by XLM-RoBERTa$_{large}$. It is trained on a larger corpus and has a larger vocabulary size. As a result, fewer unknown tokens are introduced after tokenization. In Figure \ref{fig:tranformer_tokenizers}, we present an example of a sentence tokenized using XLM-RoBERTa$_{large}$ and BERT$_{base}$. We can see that the XLM-RoBERTa$_{large}$ tokenizer performs better than the BERT$_{base}$ tokenizer. 
When we use the BERT$_{base}$ tokenizer, the last two words are replaced with the unknown tokens, even though they are not rare words in Bangla. On the other hand, the first word is split into several tokens. XLM-RoBERTa$_{large}$ performs better in this regard due to the availability of more extensive vocabulary for Bangla in the model. As XLM-RoBERTa$_{large}$ is a model trained with more layers and parameters, it also has better generalization ability compared to other models. This results in better performance in the downstream tasks.

\begin{figure}
\centering
\begin{subfigure}{\columnwidth} 
\centering
    \includegraphics[width=0.5\columnwidth]{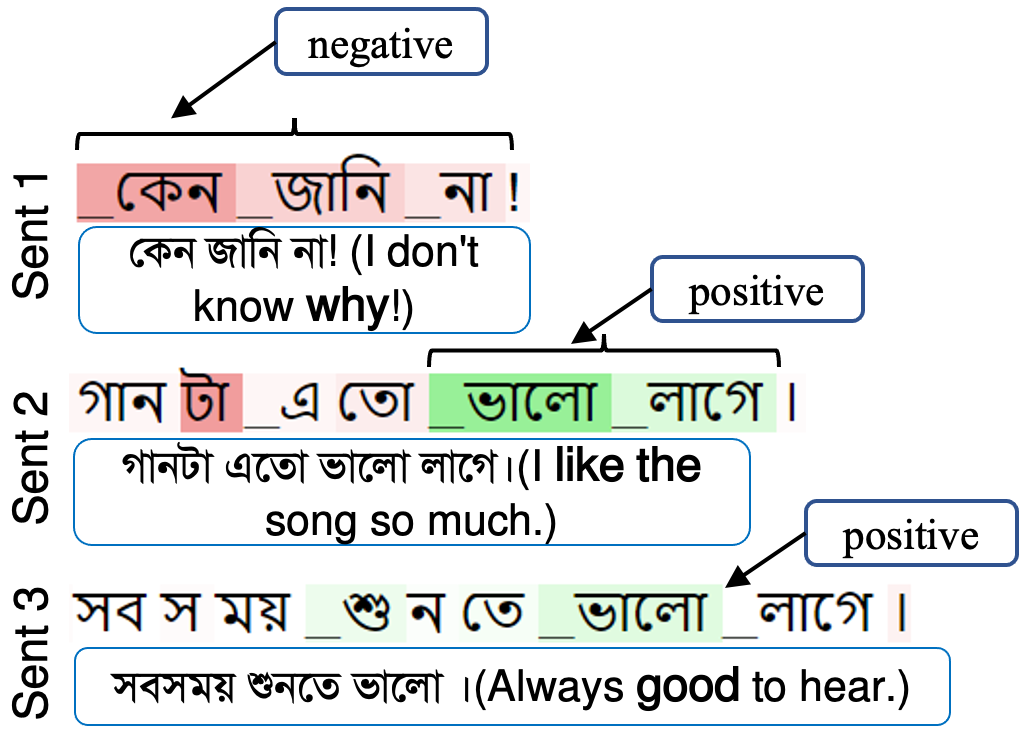}
    \caption{An example, which is detected as \textbf{positive sentiment} with a probability of 0.997.}
    \label{fig:sent1}  
\end{subfigure}
    \hfill
\begin{subfigure}{\columnwidth} 
    \centering
    \includegraphics[width=0.5\columnwidth]{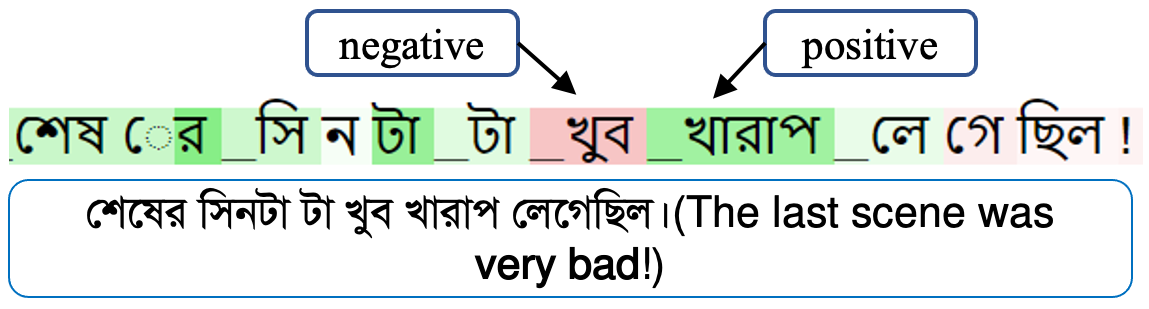}
    \caption{An example, which is detected as \textbf{negative sentiment} with a probability of 0.963.}
    \label{fig:sent2}  
\end{subfigure}
\hfill
\begin{subfigure}{\columnwidth}
    \centering
    \includegraphics[width=0.5\columnwidth]{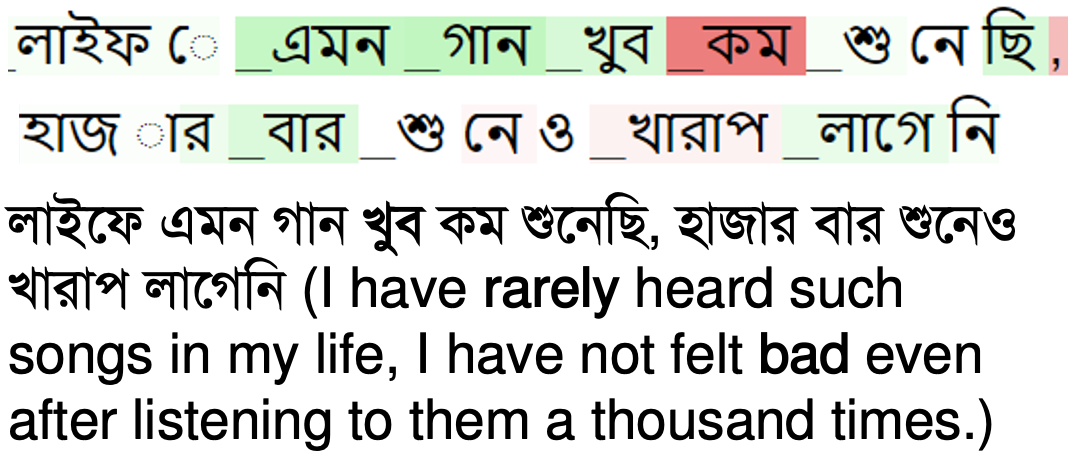}
    \caption{An example, which is detected as \textbf{positive sentiment} with a probability of 0.980.}
    \label{fig:sent3}  
\end{subfigure}
\caption{Examples from the XLM-RoBERTa$_{large}$ sentiment model highlighting the importance of words that the network learns.}
\label{fig:example_all}
\end{figure}

 The transformer-based model uses subword embedding \cite{sennrich-etal-2016-neural}. One advantage of using subword embeddings is that it can perform better in noisy user-generated text like those found in online media. As they contain misspellings and shortened words, subword embeddings (also character embedding) can still capture meaningful information. People also often comment in both Bangla and English on such platforms, as seen in the YouTube datasets. Multilingual transformer models are better suited for such types of social media text. Note that the use of transformer models for fine-tuning also reduces the need for feature engineering and data preprocessing. We did not use any separate text preprocessing steps (e.g., stop words and punctuation removal) other than using the model-specific tokenizers.

\textbf{Prediction Interpretation}: For a better interpretation of the models, we explore a sentiment model trained using XLM-RoBERTa$_{large}$ and show word importance with three examples as shown in Figure \ref{fig:example_all}. These representations are produced using Captum toolkit \cite{captum2019github}.\footnote{\url{https://github.com/pytorch/captum}} 

In Figure \ref{fig:sent1}, the model assigns a high emphasis on positive words (good). The first sentence, sent 1 (I don't know why!), can be considered to be slightly negative and is reflected in the word importance. In Figure \ref{fig:sent2},  negative words (\textbf{bad}) are properly highlighted to arrive at the overall negative sentiment. Interestingly the word (\textbf{very}) is highlighted as the opposite, i.e., positive, meaning the model was not able to combine two words and emphasize the phrase (\textbf{very bad}) as negative. Finally, the sentence in Figure \ref{fig:sent3} is more challenging. However, the model is able to identify the overall positive tone despite the presence of some negative words like \textbf{rarely}, and \textbf{bad}. From the analysis, we can conclude that transformer-based models can learn the discriminating characteristics needed for classification.




\begin{figure}[t]
\centering
\includegraphics[width=0.8\linewidth]{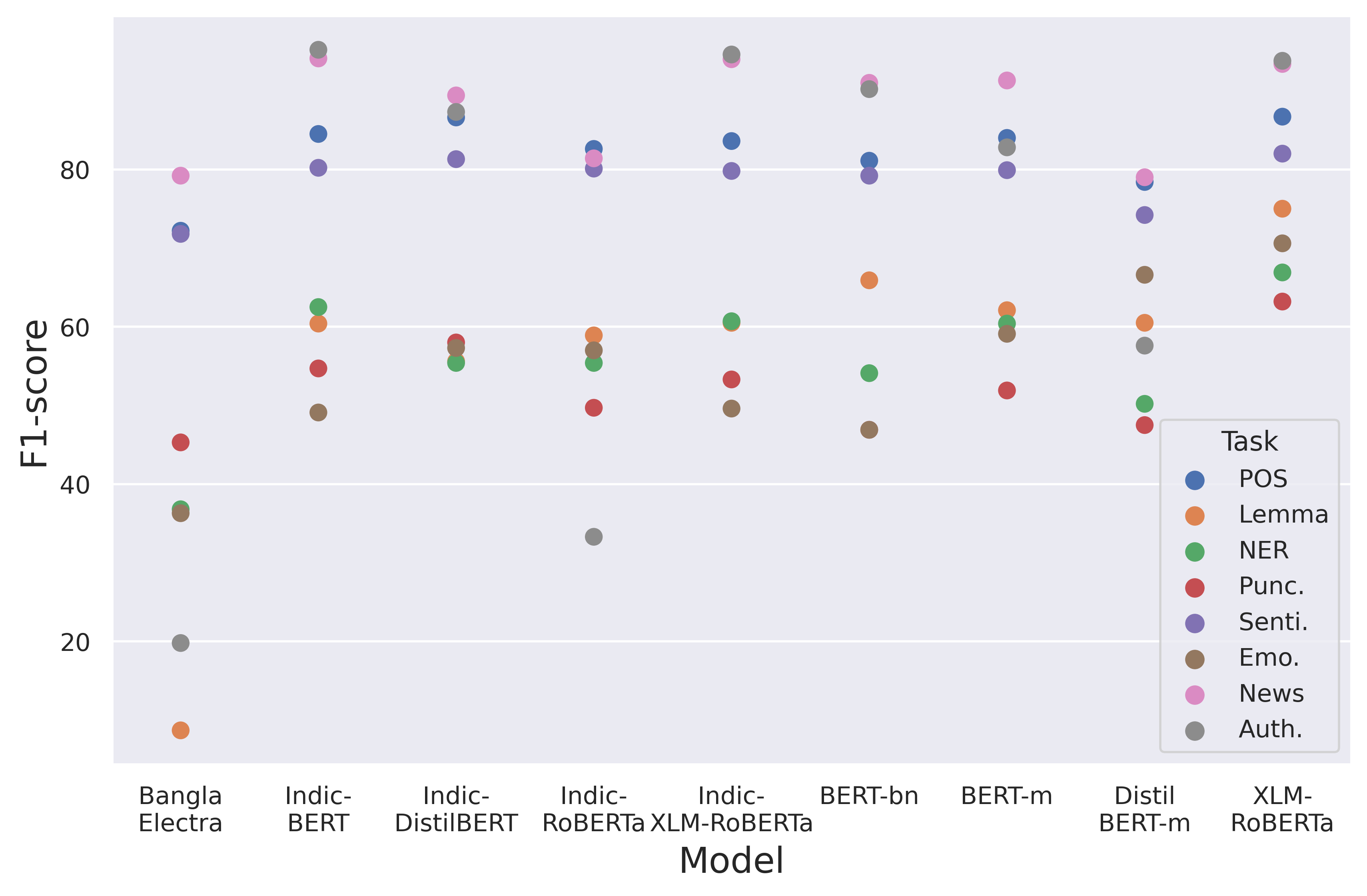}
\caption{Performance of the models across different tasks}
\label{fig:model_performance}  
\end{figure}

\begin{table}[]
\centering
\caption{Best results for different tasks.}
\label{tab:results_all_tasks}
\setlength{\tabcolsep}{4pt}
\scalebox{0.8}{
\begin{tabular}{@{}lllrrrr@{}}
\toprule
\multicolumn{1}{c}{\textbf{Task}} & \multicolumn{1}{c}{\textbf{Model}} & \multicolumn{1}{c}{\textbf{Dataset}} & \multicolumn{1}{c}{\textbf{Acc}} & \multicolumn{1}{c}{\textbf{P}} & \multicolumn{1}{c}{\textbf{R}} & \multicolumn{1}{c}{\textbf{F1}} \\ \midrule
POS & XLM-RoBERTa & \begin{tabular}[c]{@{}l@{}}LDC \cite{iitgpgpostagging}, IITKPG \cite{iitgpgpostagging}, CRBLP \cite{ummi2008developing} \cite{alam2016bidirectional}\end{tabular} & 90.1 & 87.0 & 86.4 & 86.7 \\
Lemma & XLM-RoBERTa & Chakrabarty et al. \cite{chakrabarty2017context} & 75.1 & 75.1 & 74.9 & 75.0 \\
NER & XLM-RoBERTa & CogniData \cite{chowdhury2018towards} & 87.0 & 63.6 & 70.5 & 66.9 \\
\multirow{3}{*}{Punc.} & XLM-RoBERTa & News \cite{alam-etal-2020-punctuation} & - & 87.8 & 86.2 & 87.0 \\
 & XLM-RoBERTa & Ref. Trans. \cite{alam-etal-2020-punctuation} & - & 67.6 & 70.2 & 68.8 \\
 & XLM-RoBERTa & ASR Trans. \cite{alam-etal-2020-punctuation} & - & 60.3 & 66.4 & 63.2 \\
Sentiment & XLM-RoBERTa & Combined data set \cite{AridSentiment2020} & 82.3 & 82.3 & 82.3 & 82.2 \\
Emotion & XLM-RoBERTa & Youtube comments \cite{tripto2018detecting,alam2020bangla} & 72.7 & 70.3 & 72.7 & 70.6 \\
News Cat. & Indic-BERT & News Corpus \cite{kunchukuttan2020ai4bharat,alam2020bangla} & 94.1 & 94.1 & 94.1 & 94.1 \\
Authorship & Indic-BERT & Bangla e-books \cite{khatun2019authorship,alam2020bangla} & 95.2 & 95.3 & 95.2 & 95.2 \\ \midrule
 &  &  & \multicolumn{1}{l}{\textbf{BLEU}} & \multicolumn{1}{l}{} & \multicolumn{1}{l}{} & \multicolumn{1}{l}{} \\ \midrule
MT & Transformer & Combined data set \cite{hasan2019neural} & 22.7 &  &  &  \\ \bottomrule
\end{tabular}

}
\end{table}

\subsection{Results Summary}
From our cross-architectural performance comparison, (see Figure \ref{fig:model_performance}), we observed that the XLM-RoBERTa model consistently outperforms other fine-tuned models across the tasks. We noticed Indic-BERT and Indic-XLM-RoBERTa models are competitive to each other. 
Whereas Indic-DistilBERT performs slightly worse than these two models while having less than half the number of parameters. Moreover, we noticed fine-tuned multilingual models gives better performance than the monolingual models. 
Thus reflecting the potential of utilizing the existing pretrained models to push the boundaries in BNLP and point out a need for better monolingual pretrained models.

Further analyzing cross-task performances (in Table \ref{tab:results_all_tasks}), we noticed that current performance is bound with the community engagement.
The tasks that have received wide attention in the research community have significantly higher performance. 
For example, while considering the POS tagging and punctuation restoration task, we noticed a huge difference in their current performance. 
There has been a significant amount of work for Bangla POS tagging (can be seen in Table \ref{tab:pos-table} 15 literature) which resulted to the development of reliable resources, hence better performance (POS:86.7). Whereas, we found only one study for the punctuation restoration in Bangla content, resulting in fewer resources and design choices to explore and compare performance with.  

\subsection{Future Research Directions}
There are various avenues of research that we foresee for BNLP. 

\paragraph{Reliable Resource Design:} One of the major directions for BNLP would be developing reliable resources, by following well-defined dataset development guidelines \cite{gebru2018datasheets,bender-friedman-2018-data,holland2018dataset,Hutchinson2021}. To increase reliability and ensure the quality of the annotation, we encourage the community to provide: \textit{(i)} annotation guidelines, \textit{(ii)} annotation agreement, \textit{(iii)} demographic details of the annotators. 

\paragraph{Accessible Resource:} Proper licensing and clearly providing that information can help the community to (re-)use the dataset accordingly, while giving the proper credit to the data owner. 

\paragraph{Reliable and Reproducible Benchmark:} The growth of a scientific field is often the result of the cumulative effort of the community. The research findings become hard to re-use and reproduce in the absence of a proper description of methodology, input processing, architectural parameters, proper data splits and potentially missing information of the computational environment used to carry the experiments. Besides limiting the re-usability, such a lack of information also hinders the reliability of the presented results.  
Therefore, we urge the community to collaborate in the endeavor to enrich Bangla language computing by focusing on creating reproducible work and benchmarks.


\paragraph{Strengthen Monolingual Pretrained Transformer Models:}
Efficacy of monolingual pretrained transformers are often observed in high-resource languages over multilingual counterparts \cite{martin2019camembert,nguyen2020phobert,canete2020spanish, polignano2019alberto, chowdhury2020improving}. However, such effectiveness depends on having trained the model with large and diverse training data. 
In low-resource languages, like Bangla, developing such monolingual transformer models can play a crucial part in
pushing the performance further. We believe such large monolingual models have the potential to outperform\footnote{As indicated in \cite{bhattacharjee2021banglabert}.} our results reported using XLM-RoBERTa. Thus making it an important future research avenue.


\paragraph{Technology Transfer}
Transferring applied research to a usable product is another important challenge. 
As a community, we need to engage with the industry to understand the proper technology scope, answering ``what technology to transfer'' and how to find a trade-off between the computational complexity and the model performance.

\paragraph{Broaden the Horizon}
Moreover, we also invite the community to explore several other topics that require wide attention, specifically, tasks involved in speech processing (e.g., speech recognition, speech synthesis), image processing (e.g., OCR). 

%% file: sections/conclusions.tex
\section{Conclusion}
\label{sec:conclusion}

In an effort to further enrich the Bangla natural language processing community and to keep pace with the rapidly evolving ML techniques, 
we presented our study in two folds.
First, we provide a comprehensive survey of nine different BNLP tasks through an extensive literature review. Second, we further augment the surveyed fields with our own experiments using current state-of-the-art algorithms. 
For the current study, we focused on token and text classification -- ranging from POS tagging to emotion classification, as well as 
machine translation. We reviewed 108 papers covering decades of research, and reported the resources that have been developed so far, along with approaches proposed in various studies for the aforesaid tasks. We then performed extensive experiments using available resources and the current state-of-the-art transformer models - a popular and de facto choice for text modeling.
We explored various monolingual and multilingual pretrained transformers, varying in terms of computation complexities. We performed a total of 175 set of experiments for nine different tasks using these models. Our work shows that fine-tuning the transformer models can yield better performance than the traditional methods that use hand-crafted features and other deep learning models like CNN and LSTM trained on distributed word representation. We obtained state-of-the-art results on the selected tasks, across different datasets from various domains. Our experiment results can serve as benchmarks for future work. We hope the present study will encourage researchers to make use of said models for various tasks in Bangla NLP, and serve as a stepping stone for future endeavors that will contribute to enriching BNLP research.